\RequirePackage{fix-cm}
 \documentclass[twocolumn]{svjour3}          
\smartqed  
\usepackage{graphicx}
\usepackage{amssymb}
\usepackage{amsmath,amssymb} 
\usepackage{natbib}

\newcommand{\comment}[1]{}

\def\bigO2{\mbox{${\cal O}$}}
\def\bigO{O}

\def\mA{\mathcal{A}}
\def\mB{\mathcal{B}}

\def\mH{\mathcal{H}}
\def\mK{\mathcal{K}}

\def\mN{\mathcal{N}}

\def\1n{\mathbf{1}_n}
\def\0{\mathbf{0}}
\def\1{\mathbf{1}}

\def\a{{\bf a}}

\def\a1{\mbox{\bf a}_1}
\def\a2{\mbox{\bf a}_2}
\def\a3{\mbox{\bf a}_3}
\def\a4{\mbox{\bf a}_4}

\newcommand{\ie}{\emph{i.e.\;}}
\newcommand{\eg}{\emph{e.g.\;}}

\begin{document}

\title{SEEDS: Superpixels Extracted via Energy-Driven Sampling
}


\author{Michael Van den Bergh%
\and Xavier Boix \and Gemma Roig \and
Luc Van Gool
}


\institute{M. Van den Bergh and X. Boix and G. Roig and L. Van Gool \at
              ETH Zurich - Computer Vision Laboratory\\
Sternwartstrasse 7
CH - 8092 Zurich
Switzerland
 \\
              Tel.: 	+41 44 632 52 83\\
              Fax: +41 44 632 11 99\\
              \email{\{vandenbergh,boxavier,gemmar\}@vision.ee.ethz.ch
vangool@vision.ee.ethz.ch}           
}

\date{Received: Dec 21, 2012 / Accepted: in review}

\maketitle

\begin{abstract}
Superpixel algorithms aim to over-segment the image by grouping pixels that belong to the same object.
Many state-of-the-art superpixel algorithms rely on minimizing objective 
functions to enforce color homogeneity. The optimization is accomplished 
by sophisticated methods that progressively build the
superpixels, typically by adding cuts or growing superpixels. 
As a result, they are computationally too expensive 
for real-time applications. 
We introduce a new approach based on a simple hill-climbing optimization.
Starting from an initial superpixel partitioning, 
it continuously refines the superpixels by modifying the boundaries.
We define a robust and fast to evaluate energy function,
based on enforcing color similarity between the boundaries and the superpixel color histogram.
In a series of experiments, we show that we achieve an excellent compromise between accuracy and efficiency. 
We are able to achieve a performance comparable to the state-of-the-art, but 
in real-time on a single Intel i7 CPU at 2.8GHz.
\end{abstract}

\section{Introduction}
Many computer vision applications
benefit from working with superpixels instead of just pixels~\citep[\eg][]{merdaSegmentation09,Tracking11,Objectness12, Boix11}.
Superpixels are of special interest for semantic segmentation,
in which they are reported to bring major advantages. 
They reduce the number of entities to be labeled semantically and  
enable feature computation on bigger, more meaningful regions. 

At the heart of many state-of-the-art superpixel extraction algorithms lies an 
objective function, usually in the form of a graph. The trend has been to design sophisticated 
optimization schemes adapted to the objective function, and to strike a balance between efficiency and 
performance. 
Typically, optimization methods are built upon
gradually adding cuts,
or grow superpixels starting from some estimated centers.
However, these superpixels algorithms come with a 
computational cost similar to systems producing entire semantic segmentations. 
For instance, \cite{Shotton08} report state-of-the-art
segmentation within tenths of a second per image, which is as fast as state-of-the-art
algorithms for superpixel extraction alone. 
Recent superpixel extraction methods emphasize the need for 
efficiency~\citep[\eg][]{Zhang11,Entropy11}, but still their run-time is far from real-time.

In this paper, we try another way around the superpixel problem. Instead of incrementally building the superpixels by adding cuts or growing superpixels, 
we start from a complete superpixel partitioning, and we 
iteratively refine it.
The refinement is done by moving the boundaries of the superpixels, 
or equivalently, by exchanging pixels between neighboring superpixels.
We introduce an objective function that can be  maximized efficiently, and is based on enforcing homogeneity of the color distribution of the superpixels,
 plus a term that encourages smooth boundary shapes.
The optimization is based on a hill-climbing algorithm, in which
a proposed movement for refining the superpixels is accepted  if the objective function increases. 

 \begin{figure*}[t!]
\centering
{\ Adding cuts} \\
\includegraphics[width=0.6\linewidth]{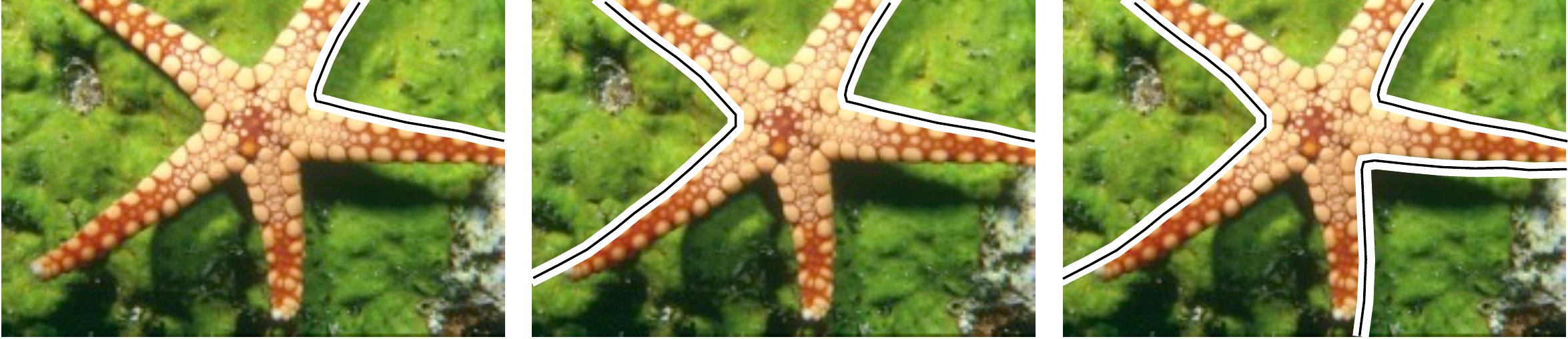} \\
\vspace{0.3cm}
{\ Growing from assigned centers\\}
\includegraphics[width=0.6\linewidth]{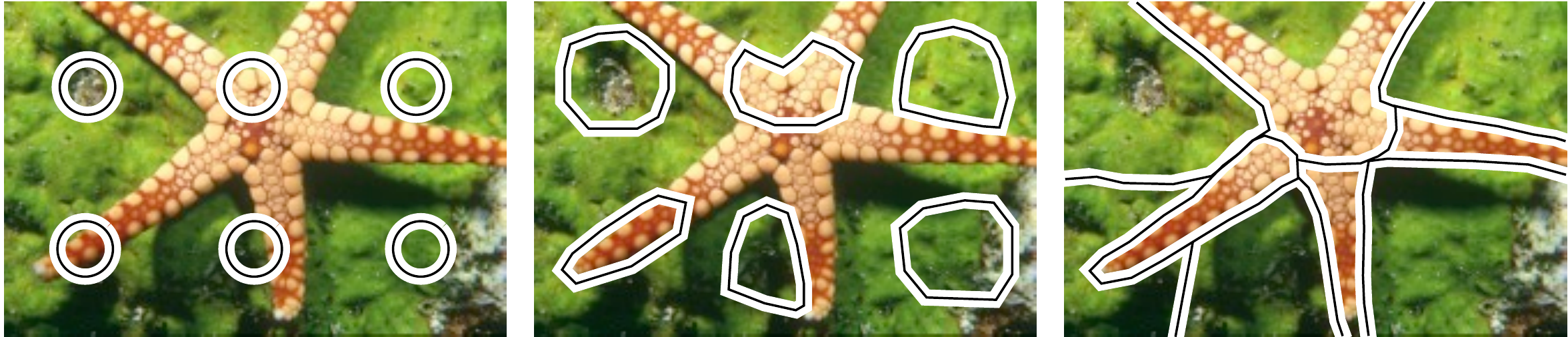}\\ 
\vspace{0.3cm}
{\ SEEDS\\}
\includegraphics[width=0.6\linewidth]{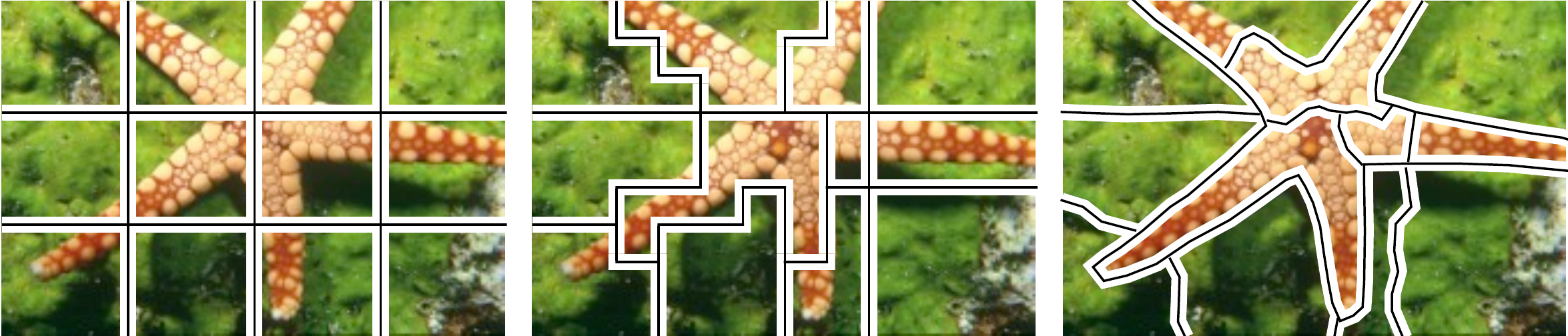}
 \caption{Comparison of different strategies to build superpixels. Top: the image is progressively cut; Middle: the superpixels grow
from assigned centers. Bottom: the presented method (SEEDS) proposes a novel approach: 
it initializes the superpixels in a gird, and continuously exchanges pixels on the boundaries between neighboring superpixels.
}
\label{aa}
\end{figure*}

We show that the hill-climbing needs few operations to evaluate the energy function.
We introduce a boundary updating using block sizes defined in a hierarchy. 
Accordingly, the boundary updating has been adapted to start with large blocks and then decreasing the block size as the algorithm iterates down to pixel-level. 
We will show this efficient exchange of pixels between superpixels enables the algorithm to run significantly faster than the state-of-the-art. 
In particular, it only requires
one memory look-up when a single pixel from the boundary is moved. 
%

We tested our approach on the Berkeley segmentation benchmark~\citep{BerkeleySegmentation01},
and propose an additional metric in order to improve the comparison with other superpixel algorithms.
We show that, to the best of our knowledge, 
the presented method (SEEDS) is faster than the fastest state-of-the-art methods and its performance is competitive with the best non-real-time methods.
Indeed, it is able to run in real-time (30Hz) using a single CPU Intel i7 at 2.8GHz without GPUs or dedicated hardware.

\section{Towards Efficiently Extracted Superpixels}
In this Section, we revisit the literature on superpixel extraction. 
The concept of superpixels as a pre-processing step was first introduced by~\cite{Ren03}. They defined the superpixels as an over-segmentation
of the image based on the principles of grouping developed by the classical Gestalt theory by~\cite{Wertheimer38}. 
We divide the existing superpixel methods in two families, putting 
special emphasis on their compromise between accuracy and run-time.  
In the first one, the methods are based on graphs and work by gradually adding cuts. 
In the other, 
they gradually grow superpixels starting from 
an initial set. 
We add a third approach, which we first introduced it in~\cite{SEEDS}, 
which moves the boundaries from an initial superpixel partitioning. 
We illustrate the different methods in Fig.~\ref{aa}.

\subsection{Gradual Addition of Cuts} 
Typically, these methods are built upon an objective function that takes the 
similarities between neighboring pixels into account and use a graph to represent 
it. Usually, the nodes of the graph represent pixels, and the 
edges their similarities. \cite{Ncuts00} introduced the seminal Normalized Cuts 
algorithm. It is based on the earlier work by~\cite{Wu93}, which
globally minimizes a graph-based objective function, 
by finding the optimal partition in the graph recursively. In \cite{Ncuts00}, the 
cut cost is improved  by normalizing it taking into account all the nodes in the graph. In this way, they avoid favouring the cuts 
in small sets of nodes in the graph. 
Normalized 
Cuts is computationally demanding, and there have been attempts 
to speed it up, by adding constraints~\citep{Eriksson07,Xu09}, or by decomposing the graph in multiple scales~\citep{Cour05}.

Another strategy to improve the efficiency of 
graph-based methods was introduced by \cite{FH04}.  
They presented an agglomerative clustering of the nodes of the graph, which is 
faster than Normalized Cuts. However, \cite{Levinshtein09} and \cite{Veksler10} showed that it 
produces superpixels of irregular size and shapes which might no be desirable. 
The algorithm by \cite{Moore08,Moore10} finds the optimal 
 cuts by using pre-computed 
boundary maps. Yet, the performance of this algorithm depends on the quality of such boundary maps. 
\cite{Veksler10} place overlapping patches over the image and assign each pixel to one of those by 
inferring a solution with graph-cuts. Based on this work, \cite{Zhang11} proposed an efficient algorithm that uses a 
pseudo-boolean optimization and achieves $0.5$ seconds per image.

Recently, \cite{Entropy11} introduced a new graph-based energy function 
and surpassed the previous results in terms of quality. Their method maximizes the entropy rate 
of the cuts in the graph, plus a balancing term that encourages superpixels of similar 
size. They show that maximizing the entropy rate favors the formation of compact and 
homogeneous superpixels, and they optimize it using a greedy algorithm.
 However, they also report that the algorithm takes about 2.5~s to segment an image 
of size $480\times 320$.



\subsection{Growing superpixels from assigned centers} 

There are methods not based on graphs. Watersheds is among the pioneers~\citep{Vincent91,Meyer99}.
It uses the gradient image, which is seen as a topological surface, and the superpixels are created by flooding
the gradient image. A more recent method based on similar principles is  Turbopixels~\citep{Levinshtein09}. It grows 
regions following geometric flows, until the superpixels are formed. 


\cite{SLIC10} introduced SLIC algorithm, which substantially improves the efficiency of superpixel extraction.
SLIC starts from a regular grid of centers or segments, and grows the superpixels by clustering pixels around the centers. 
At each iteration, the centers are updated, and the superpixels are grown again.
\cite{Zeng11} formulates this algorithm taking into account the geodesic distances between pixels, and accepts adding new superpixel centers.   
Consistent Segmentation by~\cite{zitnick} it is based on similar principles, but it also  
estimates the optical flow jointly with the segmentation in video sequences using appearance and motion constraints.

A different strategy is followed by Quick-Shift~\citep{Vedaldi08}. It performs fast mean-shift, which was introduced by~\cite{Comaniciu02}, with a 
non-parametric clustering and with a non-iterative algorithm.


Even though all these methods 
are more efficient than graph-based alternatives, they do not run in real-time, and
in most cases they obtain inferior performance. SLIC, being the fastest among them, it is able to run at $5$Hz.


\subsection{SEEDS}

Our approach is related to the methods that grow superpixels from an initial set in the sense that it also starts from 
a regular grid. Yet, it does not share their bottleneck of needing to iteratively grow
superpixels. Growing might imply computing some distance between the superpixel and all 
surrounding pixels in each iteration, which comes at a non-negligible cost.
Our method bypasses growing superpixels from a center,
 because it directly exchanges  pixels between superpixels by moving the boundaries.

\section{Superpixels as an Energy Maximization}

\begin{figure*}[t!]
\centering
\begin{tabular}{cc}
\includegraphics[width=0.25\linewidth]{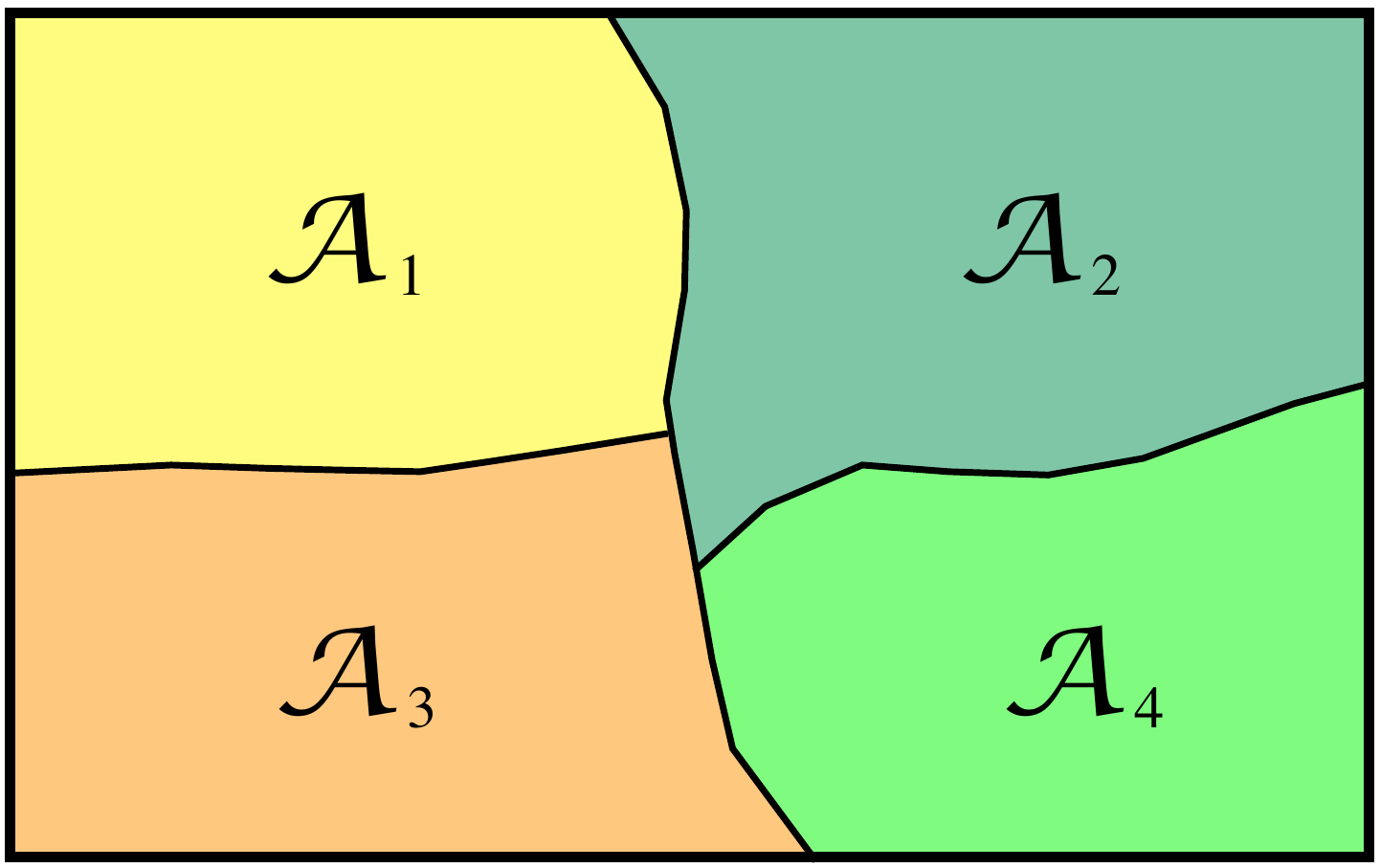} & \includegraphics[width=0.25\linewidth]{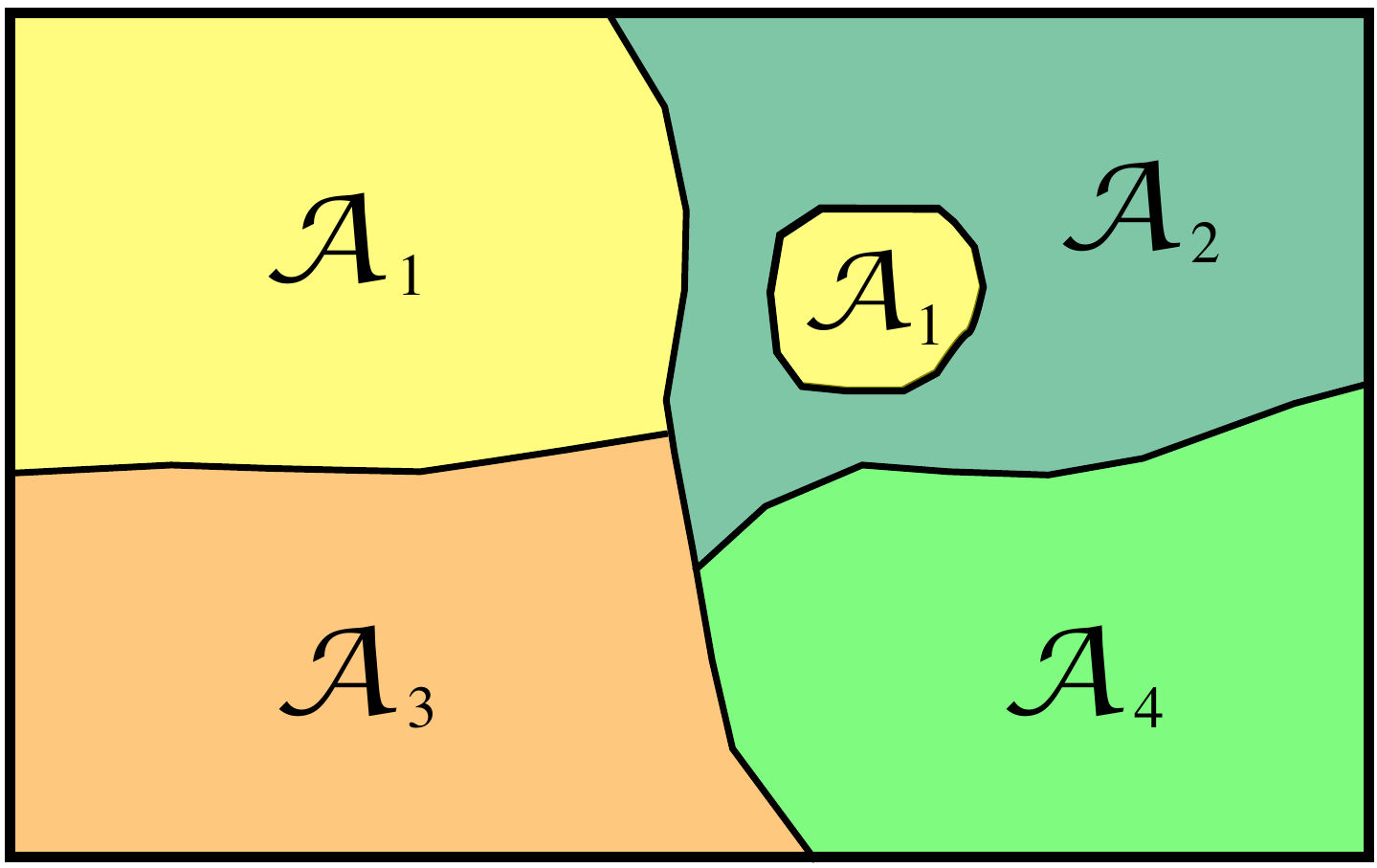} \\
\end{tabular}
 \caption{Left: an example partitioning in $\mathcal{S}$, where the 
superpixels are connected. Right: the partitioning is  in $\mathcal{C}$ but not in $\mathcal{S}$ as it is an invalid superpixel partitioning.
}
\label{figPartition}
\end{figure*}

The quality of a superpixel is measured by its property of grouping similar pixels that belong to the same object, 
and by how well it follows object boundaries. 
Therefore, a superpixel segmentation usually enforces a consistent appearance inside superpixels and a regular shape of the superpixel boundaries. 
We introduce the superpixel segmentation as an energy maximization problem where each superpixel is defined 
as a region with a color distribution and a shape of the boundary.

Let $N$ be the number of pixels in the image, and $K$ the number of
superpixels that we want to obtain\footnote{The number of desired superpixels $K$ 
is assumed to be fixed, as is usual in most previous work, which allows for 
a comparison with the state-of-the-art.}. 
We represent a partitioning of the image into superpixels  with the mapping
\begin{align}
s:\{1,\ldots,N\}\rightarrow \{1,\ldots,K\},
\end{align}
where $s(i)$ denotes the superpixel to which pixel $i$ is assigned.
Also, we can represent an image partitioning by referring to the set of pixels in a superpixel,
which we denote as $\mathcal{A}_k$: 
\begin{equation}
 \mathcal{A}_k = \{ i: s(i)=k\},
\end{equation}
and thus, $\mathcal{A}_k$ contains the pixels in superpixel $k$.
The whole partitioning of the image is represented with the sets $\{\mathcal{A}_k\}$.
Since a pixel can only be assigned to a single superpixel, 
all sets $\mathcal{A}_k$ are restricted to be disjoint, and thus,
 the intersection between any pair of superpixels is always the empty set:   $\mathcal{A}_k\cap\mathcal{A}_{k^\prime}=\emptyset$. 
In the sequel, we interchangeably use  $s$ or $\{\mathcal{A}_k\}$ to represent a partitioning of the image into superpixels.

A superpixel is valid if spatially connected as an individual blob.
We define $\mathcal{S}$ as the set of all partitionings into valid superpixels, 
and $\bar{\mathcal{S}}$ as the set of invalid partitionings, as shown in  
 Fig.~\ref{figPartition}.
Also, we denote $\mathcal{C}$ as the more general set 
 that includes all possible partitions (valid and invalid).
 
The superpixel problem  aims at finding the partitioning $s\in \mathcal{S}$ 
that maximizes an objective function, or so called energy function. We denote the energy function as $E(s,I)$, 
where $I$ is the input image. In the following, we will omit the dependency of the energy function on $I$ for simplicity of notation.
Then, we define $s^\star$ as the partitioning that maximizes the energy function:
\begin{equation}
 s^\star=\arg\max_{s \in \mathcal{S}}E(s).
\label{eqProgressiveProblem}
\end{equation}
This optimization problem is challenging because the cardinalities of $\mathcal{S}$ 
and $\mathcal{C}$ are huge. In fact, $|\mathcal{C}|$ is the Stirling number of the 
second kind, which is of the order of $\frac{K^n}{K!}$~\citep{Sharp69}. 
What also renders the exploration of $\mathcal{S}$ 
difficult, is how $\mathcal{S}$ is embedded into $\mathcal{C}$. For each element 
in $\mathcal{S}$ there exists at least one element in $\bar{\mathcal{S}}$ which only
differs in one pixel. This means that from any valid image partitioning, we are always  
one pixel away from an invalid solution.

\section{Energy Function }
\label{secEnergy}
This section introduces the energy function that is optimized, and which is defined as the sum of two terms. One term $H(s)$ is based on the likelihood of the color of the 
superpixels, and the other term $G(s)$ is an optional prior of the shape of the superpixel boundaries. 
Thus, the energy becomes
\begin{equation}
E(s) = H(s) + \gamma G(s),
 \label{energy2term}
\end{equation}
where $\gamma$ weighs the influence of each term, and is fixed to a constant value in the experiments.

\vspace*{-0.3cm}
\subsection{Color Distribution Term: $H(s)$}
The term $H(s)$ evaluates the color distribution    
of the superpixels.
By definition, a superpixel is perceptually consistent and
should be as homogeneous in color as possible. 
Nonetheless, it is unclear which is the best mathematical way to evaluate
the homogeneity of  color in a region.
Almost each paper on superpixels in the literature introduces a new energy function to maximize,
but  none of them systematically outperforms the others.
We introduce a novel measure on the color density distribution in a superpixel,
that 
allows for efficient maximization with the hill-climbing approach.

We assume that the color distribution of each superpixel
is independent from the rest. 
We do not enforce color neighboring constraints between superpixels,
since we aim at over-segmenting the image, and it might be plausible
that two neighboring superpixels have similar colors.
This is not to say that the neighboring constraints are not useful in principle,
but  our results suggest that without them we can still achieve
 excellent performance. 

\begin{figure*}[t!]
\begin{minipage}[b]{0.25\linewidth}
 \hspace*{0.5cm} \includegraphics[width=1.0\textwidth]{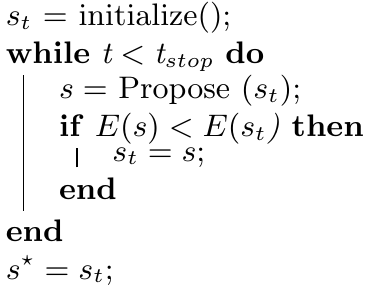} 
\end{minipage}
\hspace{1.5cm}
\begin{minipage}[t]{0.6\linewidth}
\centering
\includegraphics[width=0.9\textwidth]{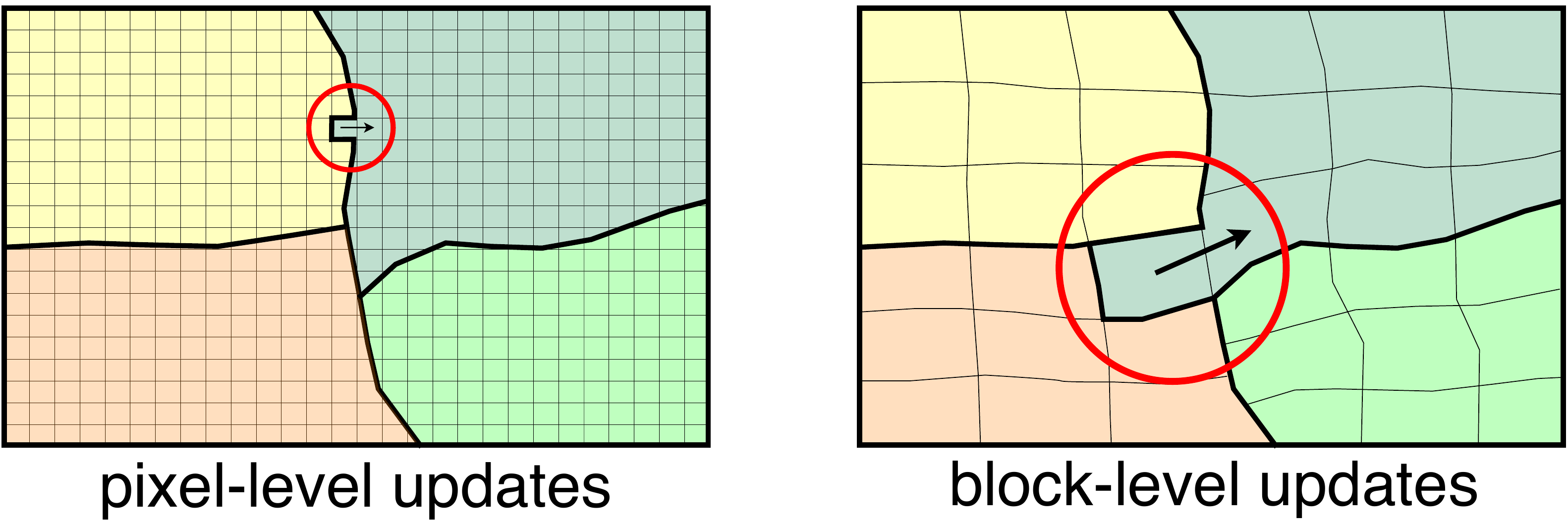} 
\end{minipage}
 \caption{Left: algorithm. Right: movements at pixel-level and at block-level. }
\label{imMovements}
\end{figure*}

Our  energy function is built upon evaluating  the color density distribution of each superpixel. 
Let $\Psi(c_{\mathcal{A}_k})$ be a quality measure of a color distribution,
and we define $H(s)$ as an evaluation of such quality in each superpixel $k$, \ie 
\begin{equation}
H(s) = \sum_{k}  \Psi(c_{\mathcal{A}_k}).
\label{eqColorSum}
\end{equation}
$\Psi(c_{\mathcal{A}_k})$ is a function that enforces that the color distribution
is concentrated in one or few colors.
A common way to approximate a density distribution
is discretizing the space into bins and building a histogram.
Let $\lambda$ be an entry in the color space, and 
$\mathcal{H}_j$ be a closed subset of the color space.
$\mathcal{H}_j$ is a set of $\lambda$'s that defines
the colors in a bin of the histogram.
We denote $c_{\mathcal{A}_k}(j)$ as the color histogram
of the set of pixels in $\mathcal{A}_k$,  and it is
\begin{equation}
 {c}_{\mathcal{A}_k}(j)=\frac{1}{Z}\sum_{i\in \mathcal{A}_k} \delta(I(i) \in \mathcal{H}_j).
\label{eqColorHistogram}
\end{equation}
$I(i)$ denotes the color of pixel $i$, and
$Z$ is the normalization factor of the histogram. 
$\delta(\cdot)$ is the indicator function, which in this case returns $1$ when 
the color of the pixel falls in the bin $j$.

We define $\Psi(c_{\mathcal{A}_k})$ to enforce that the color histogram is concentrated in few colors. 
A  valid measure could be  the entropy of the color histogram.
Yet, we found that the following measure is advantageous:
\begin{equation}
\Psi(c_{\mathcal{A}_k})=\sum_{\{\mH_j\}} (c_{\mathcal{A}_k}(j))^2.
\label{eqColorMain}
\end{equation}
In the sequel we will show that this objective function can be optimized very efficiently by a hill-climbing algorithm, as histograms can be evaluated and updated efficiently. 
Observe that $\Psi(c_{\mathcal{A}_k})$ in Eq.~\eqref{eqColorMain} encourages homogeneous superpixels,
since the maximum of $\Psi(c_{\mathcal{A}_k})$ is reached when the histogram is concentrated in one bin, which
gives $\Psi(c_{\mathcal{A}_k})=1$. In all the other cases, the function is lower,
and it reaches its minimum in case that all color bins take the same value. 
The main drawback of this energy function is that it does not take into account whether the colors are placed in bins 
far apart in the histogram or not. However, this is alleviated by the fact that we aim at over-segmenting the image, and each
superpixel might tend to cover an area with a single color.

\subsection{Boundary Term: $G(s)$}
\label{boundaryterm}

The term $G(s)$ evaluates the shape of the superpixel.
We call it boundary term and it penalizes
local irregularities in the superpixel boundaries.
Depending on the application, this term can be chosen to enforce different superpixel shapes, \eg~$G(s)$ can be chosen to favor compactness, smooth boundaries, or even proximity to  edges based on an edge map. It seems subjective which type of shape is preferred. 

Using SEEDS algorithm, we will show that this boundary term becomes optional. 
If one desires more control over the shape of the superpixels, this 
can be done inside the SEEDS framework using this boundary term $G(s)$.

In that case $G(s)$ can be defined as a local smoothness term.
Our boundary term places a $N\times N$ patch around each pixel in the image.
 Let $\mathcal{N}_i$ be the patch around pixel $i$, \ie the set of pixels that are in a squared area of size $N\times N$ around pixel $i$.
In analogy to the 
color distribution term, we use a quality measure based on a histogram. 
Each patch counts the number of different superpixels present in a local neighborhood.
We define the
histogram of superpixel labels in the area  $\mathcal{N}_i$ as
\begin{equation}
 {b}_{\mathcal{N}_i}(k)=\frac{1}{Z}\sum_{j\in \mathcal{N}_i} \delta(j \in \mathcal{A}_k).
\label{boundaryterm}
\end{equation}
Note that this histogram has $K$ bins, and each bin corresponds to a superpixel label. 
The histogram counts the amount of pixels from superpixel $k$ in the patch.

Near the boundaries, the pixels of a patch can belong to several superpixels, and 
away from the boundaries they  belong to one unique
superpixel. We consider that a  superpixel has a better shape when most of the patches 
contain pixels from one unique superpixel. We define $G(s)$ using the same measure of quality as in $H(s)$, because, as we will show, it yields an efficient optimization algorithm. Thus, it becomes
\begin{equation}
G(s) =  \sum_{i} \sum_k ({b}_{\mathcal{N}_i}(k))^2.
\label{eqBoundaryMain}
\end{equation}
If the patch $\mathcal{N}_i$ contains a unique superpixel, $G(s)$ is at its maximum. 
Observe that it is not possible that such maximum is achieved in all pixels, 
because the patches near the boundaries contain multiple superpixel labelings.
However, penalizing patches containing several superpixel labelings 
reduces the amount of pixels close to a boundary, and thus enforces regular shapes.
Furthermore, in the case that a boundary yields a shape which is not smooth, 
the amount of patches that take multiple superpixel labels is higher. A typical example to avoid is a section as thin as 1 pixel extending into neighboring superpixels. The smoothing term penalizes such cases, among others, and thus encourages a smooth labeling between superpixels.

\begin{figure*}[t!]
\centering
\includegraphics[width=1.0\textwidth]{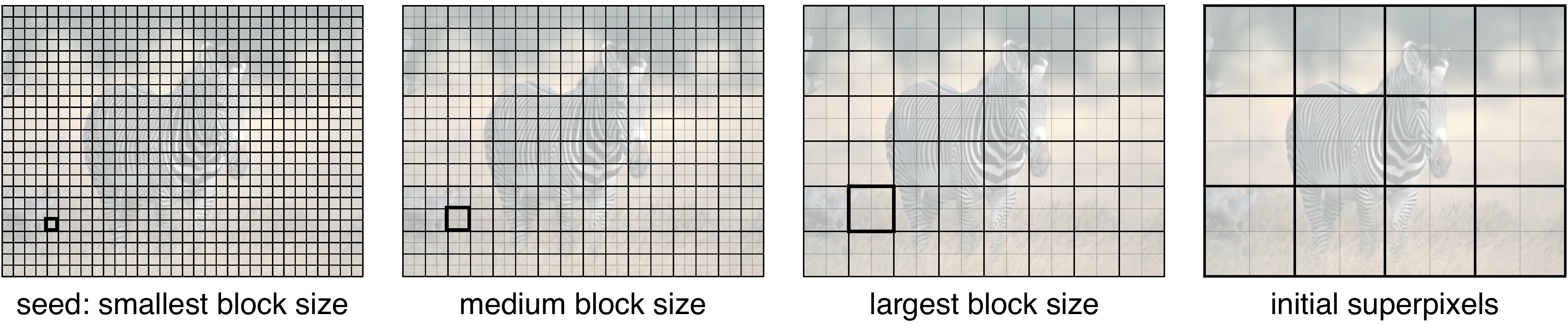}
\caption{\emph{Initialization.} Example of initialization with 12 superpixels and blocks of different sizes. 
The initialization occurs from left to right: first the smallest blocks are initialized, 
and then concatenated $2\times2$ to form larger blocks. The largest blocks are concatenated $2\times2$ 
to create the initial superpixels. This rectangular grid (in this case $4\times3$) is the starting point of the SEEDS algorithm.}
\label{blocks}
\end{figure*}

\section{Superpixels via Hill-Climbing Optimization}
\label{implement}

We introduce a  hill-climbing optimization for extracting superpixels.
Hill-climbing is an optimization algorithm that
iteratively updates the solution by proposing small local changes at each iteration. 
If the  energy function of the proposed partitioning increases, the solution is updated. 
We denote $s\in\mathcal{S}$ as the proposed partitioning, and  $s_t\in\mathcal{S}$ the lowest energy partitioning found at the instant $t$.
A new partitioning $s$ is proposed by introducing local changes at $s_t$, which in our case consists of moving some pixels from one superpixel to its neighbors. 
An iteration of the hill-climbing algorithm can be extremely efficient, because small changes to the partitioning can be evaluated very fast in practice.

An overview of the hill-climbing algorithm is shown in Fig.~\ref{imMovements}. After initialization, the algorithm 
proposes new partitionings at two levels of granularity: pixel-level and block-level. Pixel-level updates move a superpixel boundary by 1 pixel, 
while block-level updates move a block of pixels from one superpixel to another.
We will show that both types of update can be seen as the same operation, at a different scale.
Compared to our previous work in~\cite{SEEDS}, the boundary updating uses hierarchical block sizes
rather than a single block size. We show that this mechanism of block-level updating allows faster
and more accurate superpixels.

\subsection{Initialization}
\label{sec_init}

In hill-climbing, in order to converge to a solution close to the global optimum ($s^\star$), 
it is important to start from a good initial partitioning. We propose a regular grid as a
first rough partitioning, which obeys the spatial constraints of the superpixels to form a partitioning in 
$\mathcal{S}$.
In experiments, we found that when evaluating a 
grid against the standard evaluation metrics, the performance is respectable: the grid achieves a reasonable over-segmentation, but of course fails at recovering the object boundaries. 
Observe that object boundaries are maximally half of the grid size away from the grid boundaries.
This justifies using hill-climbing optimization for extracting superpixels, 
since the initialization is relatively close to the optimal solution.

Besides, we initialize the blocks of pixels (for the block movements) at different sizes, and compute the color histogram for each block. 
First, we generate the smallest block size, which is a block of $2\times2$ or $3\times3$ pixels. 
In order to generate larger block sizes,  the small blocks are hierarchally joined in a $2\times2$ fashion. The corresponding histograms 
can be obtained by summing the histograms of the composing blocks, as shown in Fig.~\ref{blocks}. 

The largest block size in the algorithm is a quarter of the target superpixel size.
Thus, the superpixels are initialized as the concatenation of $2\times2$ blocks of the largest block size. 
This results in superpixels of a consistent size, independent from the size of the input image. 
The desired number of superpixels can be obtained by choosing the initial block size and number of block levels accordingly.

\begin{figure*}[t!]
\centering
\includegraphics[width=1.0\textwidth]{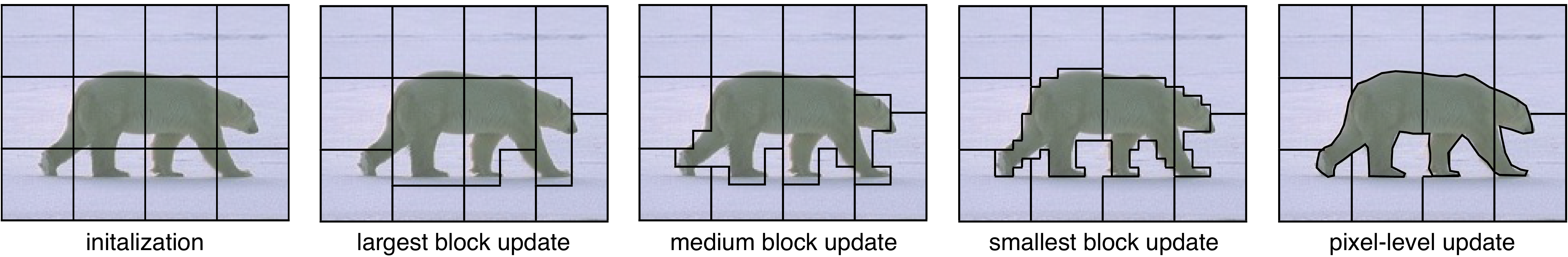}
\caption{\emph{Block and pixel movements.} This figure shows an example of the evolution of the superpixel boundaries while going through the 
iterations of the SEEDS algorithm (in the case of 12 superpixels). From left to right: 
The first image shows the initialization as a grid. The subsequent images show the block updates from large to small. 
The last image shows the pixel-level update of the superpixel boundaries.}
\label{example_iteration}
\end{figure*}

\subsection{Proposing Pixel-level and Block-level Movements}
In each iteration, the algorithm proposes a new partitioning $s$ based on the previous one $s_t$.
The elements that are changed from $s_t$ to $s$ are either single 
pixels or blocks of pixels that are moved to a neighboring superpixel.
We denote  $\mathcal{A}_k^{l}$ as a candidate set of one or more pixels to be exchanged from the superpixel $\mathcal{A}_k$ to its neighbor $\mathcal{A}_n$. 
In the case of pixel-level updates $\mathcal{A}_k^{l}$ contains one pixel (singleton), 
and in the case of block-level updates $\mathcal{A}_k^{l}$ contains a small set of pixels, as illustrated in Fig~\ref{imMovements}. 
At each iteration of the hill-climbing, we generate a new partitioning by randomly picking $\mathcal{A}_k^l$ from all boundary pixels or blocks with equal probability,
and we assign the chosen $\mathcal{A}_k^l$ to a random superpixel neighbor $\mathcal{A}_n$. In case it generates an invalid partitioning, 
which can only happen when a boundary movement  splits a superpixel in two parts, it is discarded.





Block-level updates are used for reasons of efficiency, as they allow for faster convergence, and help to avoid local maxima. 
Note that block-level updates are more expensive, but move more pixels at the same. 
Therefore, it is better to do large block-level updates at the beginning of the algorithm, 
and then smaller blocks, and finish the algorithm with pixel-level tuning of the boundaries.
Thus, we start updating at the largest block size, and
 then hierarchically move on to smaller block sizes, and 
finally the individual pixels. This is illustrated in Fig.~\ref{example_iteration}. The longer
 the individual pixel updating is run, the more accurate the resulting superpixels will be.


\subsection{Evaluating Pixel-level and Block-level Movements}
The proposed partitioning $s$ is evaluated using the energy function (Eq.~\eqref{energy2term}). 
In the following we describe the efficient evaluation of $E(s)$, and the efficient updating of the color distributions in case $s$ is accepted. 
The proofs of the propositions in this section are provided in the appendix.

\subsubsection{Color Distribution Term.} 

We introduce an efficient way to evaluate $H(s)$ 
based on the intersection distance.
Recall that the intersection distance between two histograms is
\begin{equation}
\mathbf{int}({c}_{\mathcal{A}_a},{c}_{\mathcal{A}_b}) = \sum_j
\min\{{c}_{\mathcal{A}_a}(j),{c}_{\mathcal{A}_b}(j)\},
\end{equation}
where $j$ is a bin in the histogram. Observe that it only involves
$|\{\mathcal{H}_j\}|$ comparisons and sums, where
$|\{\mathcal{H}_j\}|$ is the number of bins of the histogram.
Recall that $\mathcal{A}^l_k$ is the set of pixels that are candidates
to be moved from the superpixel $\mathcal{A}_k$ to
$\mathcal{A}_n$. We base the evaluation of $H(s)>H(s_t)$ on the
following Proposition.
\begin{proposition}
Let the sizes of $\mathcal{A}_k$ and $\mathcal{A}_n$ be similar, and
$\mathcal{A}_k^l$ much smaller, \ie $|\mA_k|\approx |\mA_n| \gg |\mA_k^l|$.
If the histogram of $\mathcal{A}^l_k$ is concentrated in a single bin,
then
\begin{equation}
 \mathbf{int}({c}_{\mathcal{A}_n}
,{c}_{\mathcal{A}_k^l})
\geq
\mathbf{int}({c}_{\mathcal{A}_k \backslash \mathcal{A}_k^l },{c}_{\mathcal{A}_k^l}) \iff
H(s) \geq H(s_t).
\end{equation}
\label{propMain}
\end{proposition}
\vspace{-0.7cm}
Proposition~\ref{propMain} can be used to evaluate whether the energy
function increases or not by simply
computing two intersection distances.
However, it makes two assumptions about the superpixels.
The first is that the size of $\mathcal{A}_k^l$
is much smaller than the size of the superpixel, and that both
superpixels have a similar size. When $\mathcal{A}_k^l$ is a single
pixel or a small block of pixels, it is
reasonable to assume that this is true for most cases.
The second assumption is that the histogram of
$\mathcal{A}_k^l$ is concentrated in a single bin. This is always the case if $\mathcal{A}_k^l$ is
a single pixel, because there is only one color. In the block-level case it is reasonable to expect that the colors in each block
are  concentrated in few bins.
In the experiments section, we show that when running the algorithm
these assumptions hold in $93\%$ of the cases.

\begin{figure}[t!]
 \centering
\includegraphics[scale=0.25]{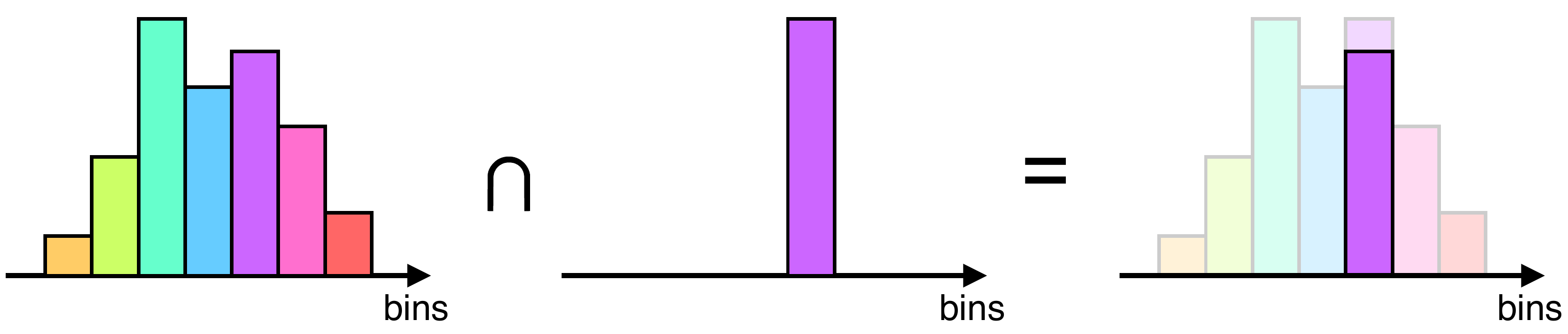}
\caption{The intersection between two histograms, when one is the color distribution of a single pixel, can be computed with a single access to memory.}
\label{imhist}
\end{figure}

Interestingly, in the case of evaluating a pixel-level update,
the computation of the intersection can be achieved with a
single access to memory, as depicted in Fig.~\ref{imhist}. This is because
the color histogram of a pixel has a single bin activated with a $1$,
and hence, the intersection distance is the value of the histogram of
the superpixel.

\subsubsection{Boundary Term.} 

The hierarchical updating of the boundaries allows us to drop the boundary term and still obtain smooth superpixel boundaries. 
This is because boundaries are updated starting with large updates and ending with fine, pixel-level updates.
Without the use of a boundary term, the energy function $E(s)$ can be evaluated more efficiently, and the method is more theoretically sound (no ad-hoc priors optimizing subjective qualities). Therefore, in the experiments section, we present the results without the use of a boundary term.
However, if one desires more control over the shape of the superpixels, this can be done inside the SEEDS framework using this boundary term $G(s)$.

During pixel-level updates, $G(s)$ can then be evaluated efficiently based on the following proposition.
%
\begin{proposition}
Let $\{b_{\mathcal{N}_i}(k)\}$ be the histograms of the superpixel
labelings computed at the partitioning $s_t$ (see Eq.~\eqref{boundaryterm}).
 ${\mathcal{A}^{l}_k}$ is a pixel, 
 and $\mathcal{K}_{\mathcal{A}^{l}_k}$ the set of pixels whose patch intersects
with that pixel, \ie $\mathcal{K}_{\mathcal{A}^{l}_k}=\{i : {\mathcal{A}^{l}_k}\in \mathcal{N}_i \}$.
If the hill-climbing proposes moving a pixel ${\mathcal{A}^{l}_k}$
from superpixel $k$ to superpixel $n$,
then
\begin{equation}
\sum_{i\in \mathcal{K}_{\mathcal{A}^{l}_k}} \left( b_{\mathcal{N}_i}(n) + 1\right)
\geq \sum_{i\in\mathcal{K}_{\mathcal{A}^{l}_k}} b_{\mathcal{N}_i}(k)\iff
G(s)\geq G(s_t).
\end{equation} 
\label{propMain2}
\end{proposition}
\vspace{-0.5cm}
Proposition~\ref{propMain2} shows that the difference in $G(s)$ can be evaluated with just a few sums of integers.

Note that Proposition~\ref{propMain2} is for pixel-level movements. 
In case of block-level updates, when assigning a block to a new superpixel, a small irregularity might be introduced at the junctions. 
Yet, note that the block boundaries are fixed unless they coincide with a superpixel boundary, in which case they can be updated in the pixel-level updates.
Smoothing these out requires pixel-level movements, thus they are smoothed in subsequent pixel-level iterations of the algorithm.

%
\begin{figure*}[t!]
\centering
\includegraphics[width=0.6\linewidth]{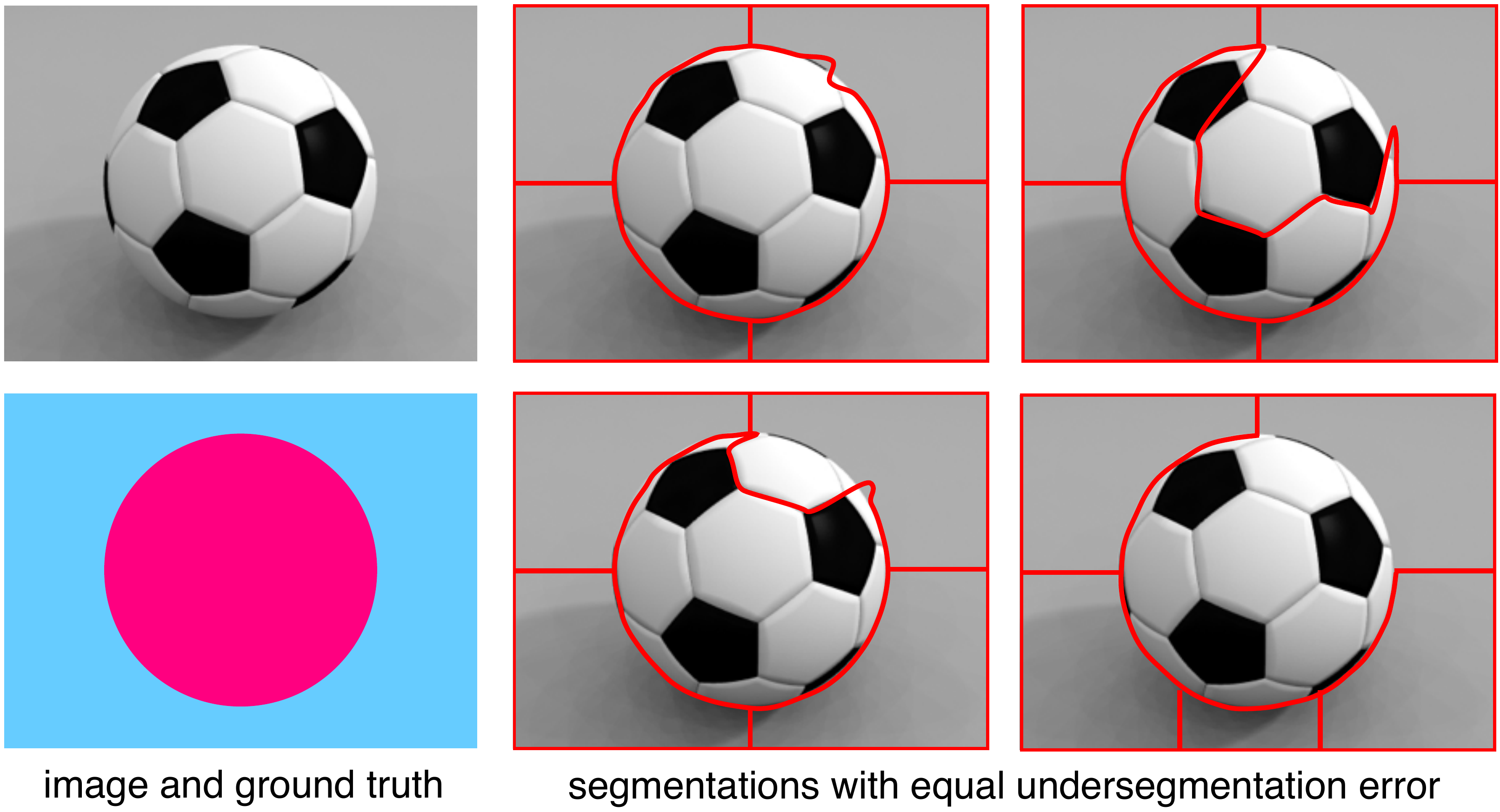}
\caption{Example of segmenting an image with 5 superpixels. In all 4 of the cases, the undersegmentation error is equal (the area of the ball and the upper right quadrant divided by the total area of the image). Even though the quality of the segmentation of the first segmentation is clearly better, it is penalized equally to the other examples.}
\label{figUE}
\end{figure*}

\subsubsection{Updating the Color Distributions.}
Once a new partition has been accepted, the histograms of $\mathcal{A}_k$ and $\mathcal{A}_n$ have to be updated efficiently. In the pixel-level case, this update can be achieved with a single increment and decrement of bin $j$ of the the respective histograms. In the block-level case, this update is achieved by subtracting $c_{\mathcal{A}^{l}_k}$ from $c_{\mathcal{A}_k}$ and adding it to $c_{\mathcal{A}_n}$.

\subsection{Termination}

When stopping the
algorithm, one obtains a valid image partitioning with a quality depending on
the allowed run-time.
The longer the algorithm is allowed to run, the 
higher the value of the objective function will get. The algorithm will usually be terminated during pixel-level updating of the boundaries. However, should one choose to terminate the algorithm very early on in the algorithm during the block-level updates, the algorithm still returns a valid partitioning.

We can set $t_{stop}$ depending on
the application, or we can even assign a time budget \emph{on the fly}.
We believe this to be a crucial property for on-line 
applications, but nonetheless one that has received little attention in the context 
of superpixel extraction so far.  
In graph-based superpixel algorithms, one has to wait until all cuts have been added to the graph,
and in methods that grow superpixels, 
one has to wait until the growing is done, the cost of which is not negligible.
The hill-climbing approach uses a lot more iterations than previous methods, but each iteration is done extremely fast.
This enables stopping the algorithm at any given time, because the time to finish the current iteration
is negligible.

\section{Experiments}

We report results on the Berkeley Segmentation Dataset (BSD)~\citep{BerkeleySegmentation01},
using the standard metrics to evaluate superpixels, as used in most recent superpixel papers~\citep{Entropy11,SLIC10,Veksler10,Levinshtein09,Zeng11}. 
We also propose a new metric for completeness and further evaluation of superpixels.
The BSD consists of $500$ images split into $200$ training, $100$ validation 
and $200$ test images. 
We use the training images to set the only parameter that needs to be tuned, and report the results based on the $200$ test images.
We compare SEEDS to defined baselines 
and to the current state-of-the-art methods.  All experiments are done using a single CPU (2.8GHz i7). We do not use any parallelization, GPU or dedicated hardware.

\subsection{Metrics}
We compute the standard metrics used to evaluate the performance of superpixel algorithms,
which are undersegmentation error (UE), boundary recall (BR) and achievable segmentation accuracy (ASA).
Additionally, we introduce a new metric, which is a corrected undersegmentation error (CUE). For UE and CUE, the lower the better, 
and for BR and ASA the higher the better. 
For completeness we also report the precision-recall curves for the contour detection benchmark proposed by~\cite{malik2011}. 
This countour benchmark allows for an evaluation of the boundary performance of the different superpixel algorithms.


\subsubsection{Undersegmentation Error (UE)}

 The undersegmentation error measures that a superpixel should not overlap
 more than one object. The standard formulation is
\begin{equation}
 UE(s)=
\frac{\sum_i\sum_{k:s_k\cap g_i\neq  \emptyset}|s_k-g_i|}{\sum_i|g_i|}
\label{ue}
\end{equation}
    where $g_i$ are the ground-truth segments, $s_k$ the output segments of the algorithm, 
    and $|a|$ indicates the size of the segment.

We found that in previous works, the 
evaluation changes slightly depending on the paper, because it is not clear in   
this measure how to treat the pixels that lie on or near a border between two labels.     
Moreover, with this metric, a segmentation based on a rectangular grid outperforms SLIC superpixels~\citep{SLIC10} and the superpixels from~\cite{FH04} (see Fig.~\ref{imevalBSD}).

In Eq.~\eqref{ue}, a single pixel error along the boundary of an object will fully penalize the superpixel it belongs on both sides of the boundary. This is illustrated in Fig.~\ref{figUE}.
Since object boundaries lie between pixels and not on pixels, this type of error can occur often. To circumvent this problem, most previous superpixel authors introduce a tolerance.                                                                                 
For instance, SLIC~\citep{SLIC10} reports a $5\%$ tolerance margin for the overlap of $s_k$ with
$g_i$; and in Entropy Rate superpixels~\citep{Entropy11} the borders of $s_k$ are removed from the
labeling before computing the UE. 
 This type of solution is rather ad hoc, and therefore, in the next section, we propose a new undersegmentation error metric, which overcomes this problem.

\begin{figure*}[t!]
\centering
\includegraphics[width=1\textwidth]{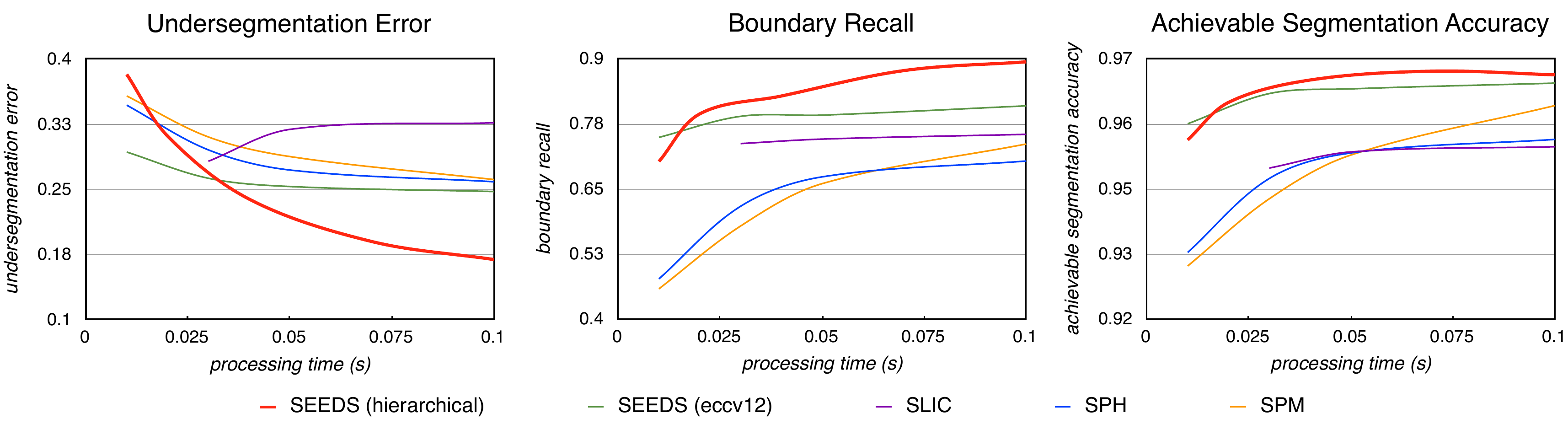}
\caption{Evaluation of SEEDS, the baselines SPH and SPM, and SLIC, versus run-time (better seen in color).}
\label{imetimeBSD}
\end{figure*}

\subsubsection{Corrected Undersegmentation Error (CUE)}

In order to compute the corrected undersegmentation error, each superpixel is matched to a single ground-truth element (largest overlap). 
Then, the number of pixels that lie outside of that ground-truth element are counted. 
This value is summed for all the superpixels and divided by the total number of pixels in the image:
\begin{equation}
 CUE(s)=
\frac{\sum_k|s_k  - g_{\max}(s_k)|}{\sum_i|g_i|},
\end{equation}
where $s_k$ are the output segments of the algorithm and $g_{\max}(s_k)$ the matching ground-truth segments with largest overlap, \ie
\begin{equation}
 g_{\max}(s_k) = \arg\max_i |s_k \cap g_i|,
\end{equation}
where $g_i$ are the ground-truth segments.

This is similar to the UE, except that the error is only counted for one side of the superpixel, not both. 
This measure will penalize the errors depending on the magnitude of the mistake. According to this measure, the errors illustrated in Fig.~\ref{figUE} will have different error.
Furthermore, it is not necessary to introduce tolerances and we believe it is a more accurate representation of the undersegmentation error.

\subsubsection{Boundary Recall (BR)}

The boundary recall evaluates the percentage of borders from the ground-truth 
    that coincide with the borders of the superpixels. It is formulated as       
\begin{equation}
 BR(s)=\frac{\sum_{p\in\mB(g)} \mbox{I}[\min_{q\in\mB(s)}\|p-q\| < \epsilon]}{|\mB(g)|},
\end{equation}
where $\mB(g)$ and $\mB(s)$ are the union sets of superpixel boundaries of the ground-truth
and the computed superpixels, respectively. The function $\mbox{I}[\cdot]$, is an indicator
 function that returns $1$ if a boundary pixel of the output superpixel is within
 a number of   pixels of tolerance, $\epsilon$, of the ground-truth boundaries. We set $\epsilon= 2$, as in~\cite{Entropy11}.

\subsubsection{Achievable Segmentation Accuracy (ASA)}

 Achievable segmentation accuracy is an upper bound measure. It gives the 
    maximum performance when taking superpixels as units for object segmentation, 
    and is computed as                 
\begin{align}
 ASA(s)=\frac{\sum_k \max_i |s_k \cap g_i|}{\sum_i |g_i|},
\end{align}
 where the superpixels are labeled with the label of the ground-truth segment
    which has the largest overlap.

       We reproduce all the results and comparisons to~\cite{SLIC10},~\cite{Entropy11} and \cite{FH04}  using the source
    code provided by the authors’ web pages. All results are computed from scratch
    using the same evaluation metrics and the same hardware across all methods.

\subsection{Parameters}

We use LAB color space, which in our experiments yields the highest performance.
%
The choice of weight $\gamma$ of $G(s)$ and size of the local neighborhood $N\times N$ is difficult 
to evaluate because there is no standard metric 
for smoothness or compactness of a superpixel in the literature. 
In fact, there is a trade-off between increasing the smoothness and the performance on the existing metrics (UE, BR and ASA). 
Therefore, in order to maximize the performance, we set $\gamma$ to 1 and $N\times N$ to the minimum size $3\times 3$. 
In the next subsection we will show the impact of the boundary term and we will compare different criterion for the boundary term. 

Since we have a variable block size and a hierarchical updating, only one parameter needs to be tuned: the number of bins in the histograms. 
This parameter is tuned on a subset of the BSD training set.
We set the number of bins to $5$ bins per color channel ($125$ bins in total), which we found to have the best performance.

We also evaluated the assumptions from Proposition~\ref{propMain} over all the updates when segmenting the training set, 
by explicitly computing the energy function in 
each iteration and comparing it to the intersection distance. 
This experiment shows that the approximation holds for $97\%$ of the pixel-level updates, and for $89\%$ of the block-level updates.



\begin{figure*}[t!]
\centering
\begin{tabular}{@{\hspace*{+0.03cm}}c@{\hspace*{0.1cm}}c@{\hspace*{0.1cm}}c@{\hspace*{0.1cm}}c}
\includegraphics[scale=0.25]{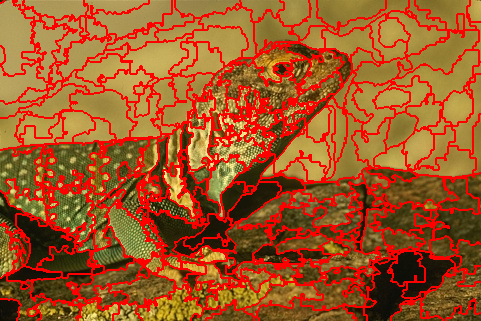} & \includegraphics[scale=0.25]{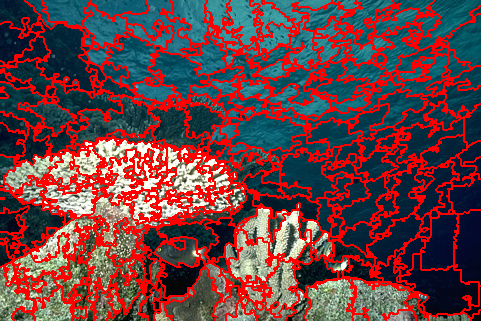} & \includegraphics[scale=0.25]{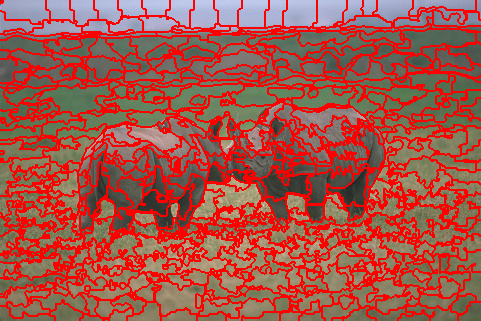} & \includegraphics[scale=0.25]{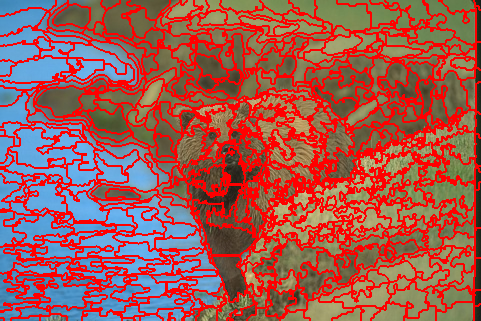}\\
\multicolumn{4}{c}{(a) SEEDS without boundary prior term}\\
\\
\includegraphics[scale=0.25]{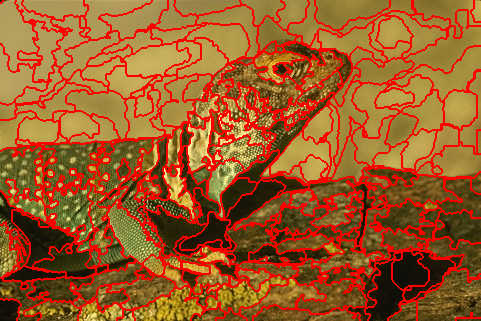} & \includegraphics[scale=0.25]{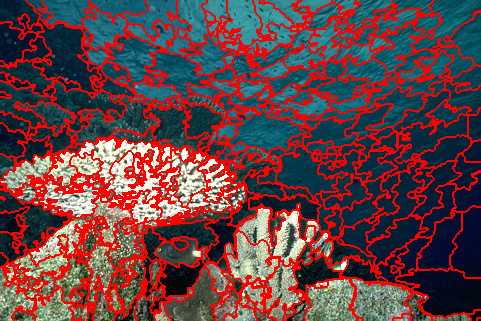} & \includegraphics[scale=0.25]{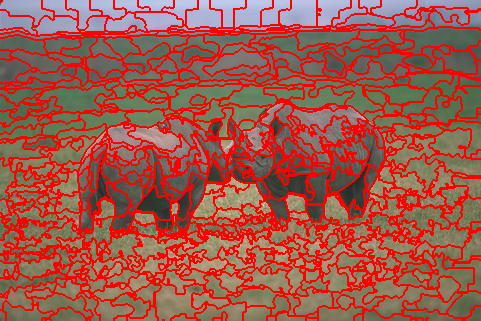} & \includegraphics[scale=0.25]{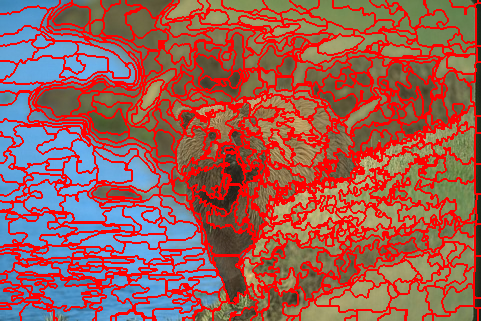}\\
\multicolumn{4}{c}{(b) SEEDS with $3\times3$ smoothing prior}\\
\\
\includegraphics[scale=0.25]{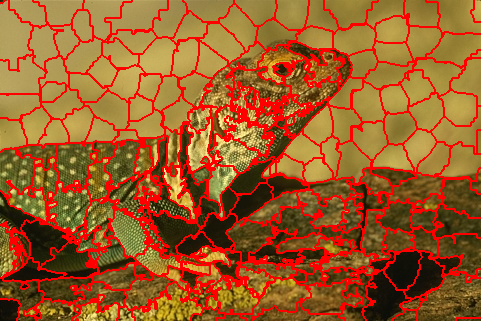} & \includegraphics[scale=0.25]{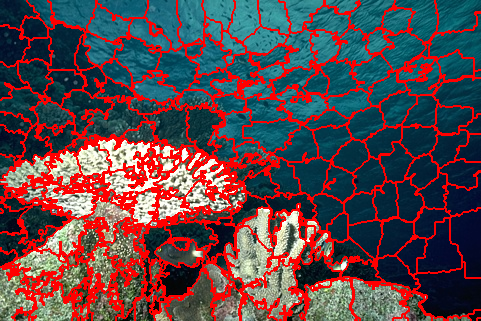} & \includegraphics[scale=0.25]{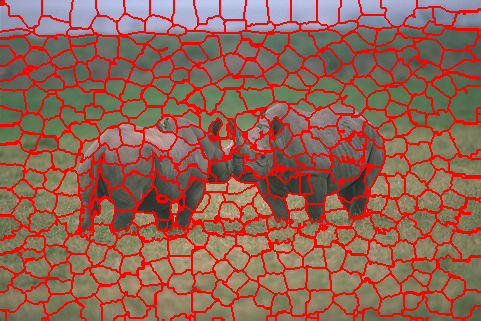} & \includegraphics[scale=0.25]{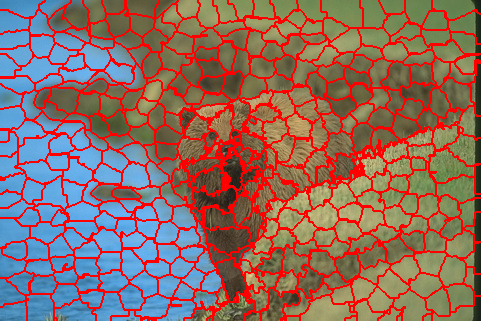}\\
\multicolumn{4}{c}{(b) SEEDS with compactness prior}\\
\\
\includegraphics[scale=0.25]{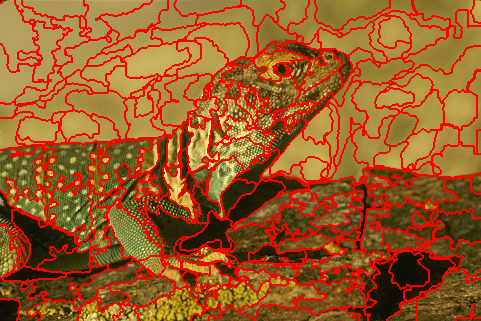} & \includegraphics[scale=0.25]{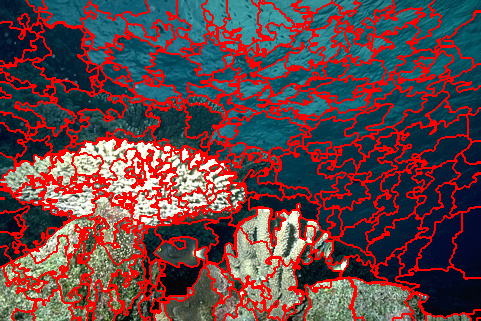} & \includegraphics[scale=0.25]{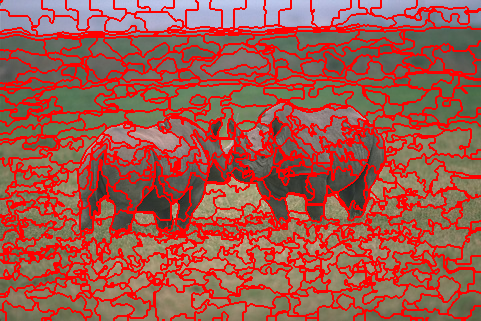} & \includegraphics[scale=0.25]{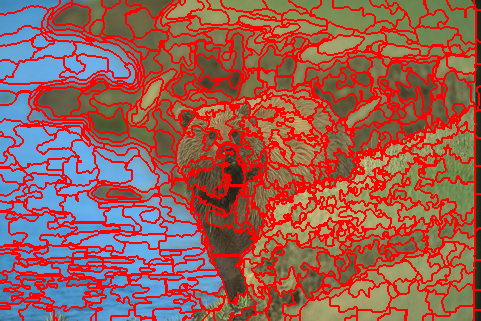}\\
\multicolumn{4}{c}{(b) SEEDS with edge prior (snap to edges)}\\
\\
\includegraphics[scale=0.25]{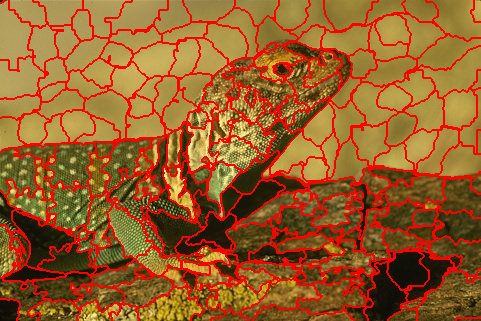} & \includegraphics[scale=0.25]{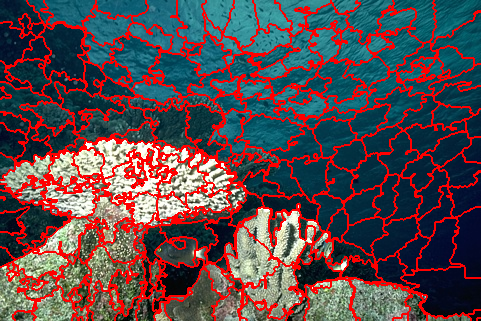} & \includegraphics[scale=0.25]{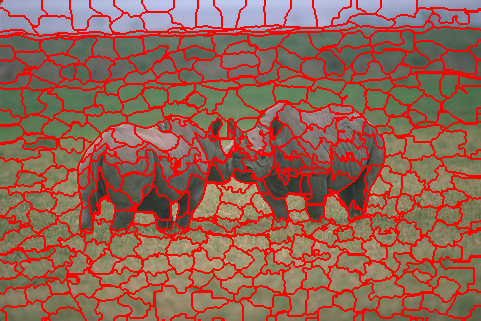} & \includegraphics[scale=0.25]{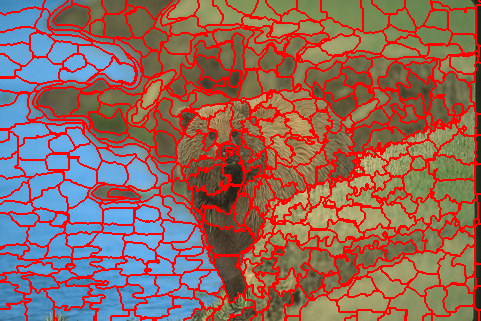}\\
\multicolumn{4}{c}{(b) SEEDS with combined prior ($3\times3$ smoothing + compactness + snap to edges)}\\
\end{tabular}
\caption{Experiment illustrating how SEEDS can produce different superpixel shapes, using the boundary prior term $G(s)$.}
\label{impriors}
\end{figure*}

\begin{figure*}[t!]
\includegraphics[width=1\textwidth]{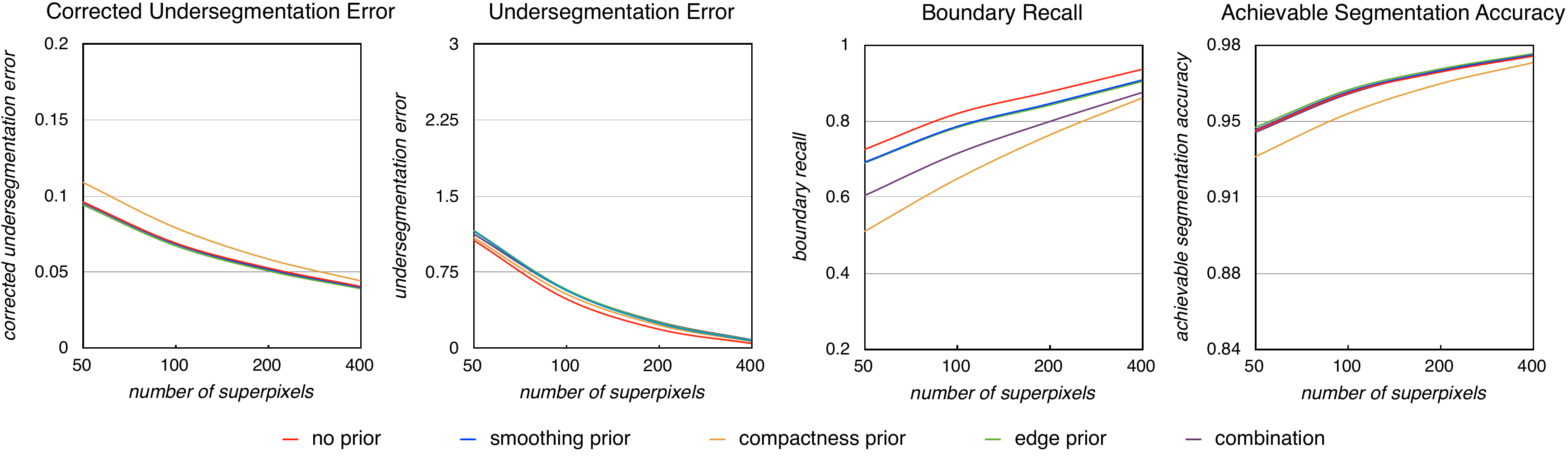}
\caption{Evaluation of SEEDS using different boundary prior terms (better seen in color).}
\label{imevalpriors}
\end{figure*}

\begin{figure*}[t!]
\includegraphics[width=1\textwidth]{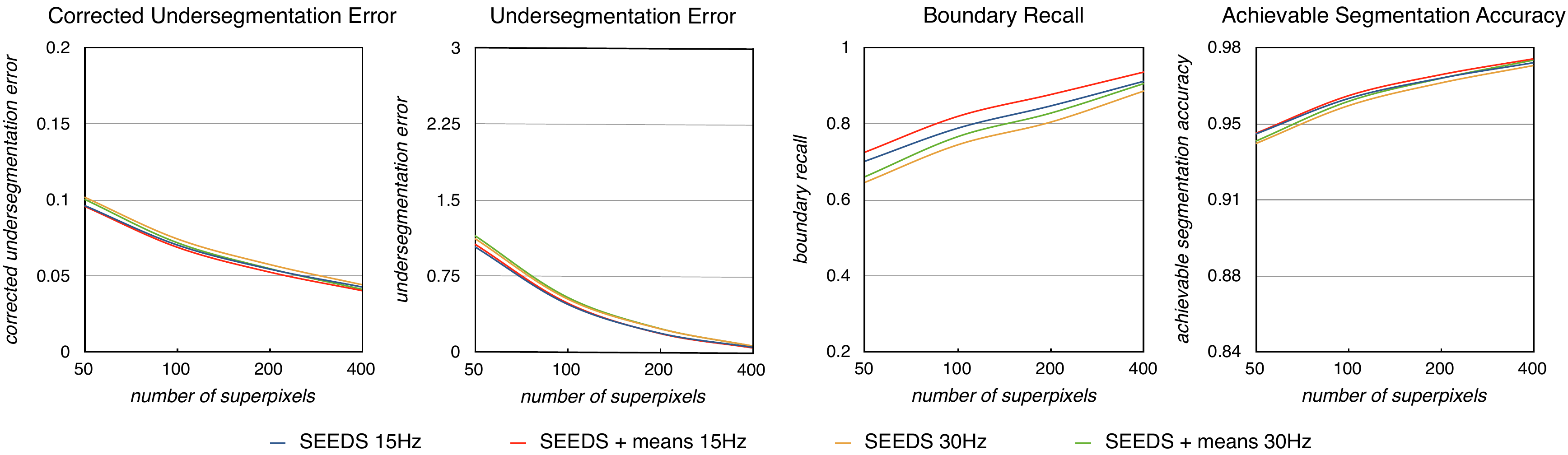}
\caption{Evaluation of SEEDS running at different speeds (15Hz and 30Hz) and with or without the means-based post-processing (better seen in color).}
\label{imevalpixel}
\end{figure*}

\begin{figure*}[t!]
\includegraphics[width=1.0\textwidth]{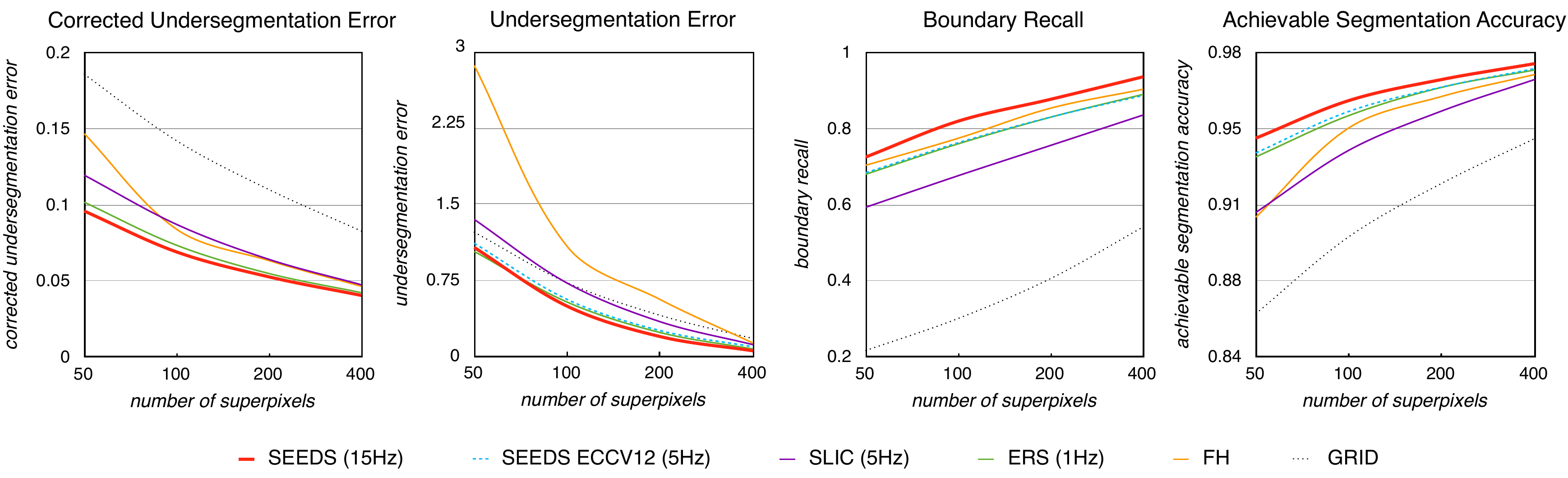}
\vspace{-0.5cm}
\caption{Evaluation of SEEDS versus the state-of-the-art on the BSD test set (better seen in color).}
\label{imevalBSD}
\end{figure*}

\subsection{Histograms and Block-level Updates}
In order to demonstrate the speed and performance benefit of block-level updates, we introduce a baseline method without block-level updates called {\em SPH} (Pixel-level using Histograms). 
This method is identical to SEEDS, except that it only uses pixel-level updating.
To demonstrate the benefit of using histograms as a color distribution, we introduce a second baseline using the mean-based distance measure from SLIC~\citep{SLIC10}, called {\em SPM} (Pixel-level using Means).

The results of this experiment are presented in function of available processing time, shown in Fig.~\ref{imetimeBSD}. The results show that SEEDS converges faster than SLIC: where SLIC requires 200 ms to compute 10 iterations, SEEDS only takes 20 ms to produce a similar result. The experiment also shows that SEEDS using histograms (SPH) converges faster than using means (SPM), and that both converge to similar results, albeit SPM slightly better. Furthermore, it shows that SEEDS converges faster when using block updates (SEEDS) than without (SPH), and to a better result, as it is less prone to getting stuck in local maxima. 
There is an anomaly where SLIC's UE seems to get worse with each iteration. We believe that this caused by SLIC's stray labels, which are only removed at the end of all iterations and might affect the performance during the iterations.

\begin{figure}[t!]
\centering
\includegraphics[width=0.3\textwidth]{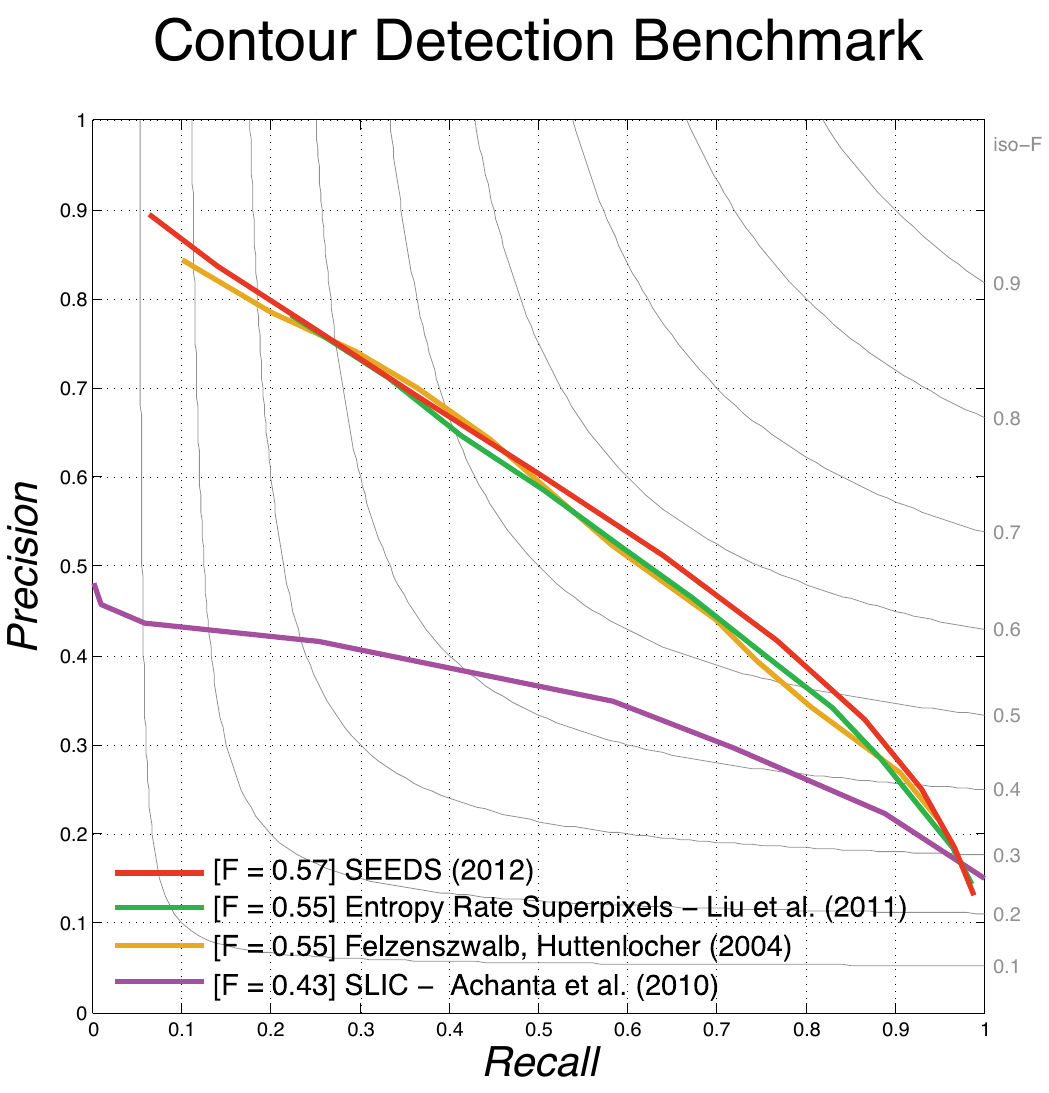}
\caption{Evaluation of SEEDS versus the state-of-the-art on the BSDS300 contour detection benchmark (better seen in color).}
\label{figCD}
\end{figure}

\subsection{Boundary Term}

In Section~\ref{boundaryterm}, we instroduced $G(s)$ as an optional boundary term. 
This prior term allows us to influence the shape of the superpixels produced by the SEEDS algorithm. 
In this section we evaluate how $G(s)$ can influence the shape of the superpixels, and how this impacts the performance. 
To this end, we compare four different prior terms. The first one is the $3\times3$ smoothing term introduced in Section~\ref{boundaryterm}. 
This is a prior which enforces local smoothing in a $3\times3$ area around the superpixel boundary. 
Second, we try a prior term based on compactness, which aims to minimize the distance between the pixels on the superpixel boundary and the center of gravity of the superpixel. 
This is similar to the compactness term in SLIC~\citep{SLIC10}, and results in superpixels that are visually similar to SLIC superpixels. Third, we introduce an edge prior. 
This is achieved by calculating a vertical and horizontal color edge map (besides the LAB color channels). 
If a boundary is near an edge, it snaps to this edge and is no longer updated from there on forward. 
If a boundary is not near an edge, it is smoothed using the $3\times3$ smoothing as described above. 
Finally, we introduce a combined prior, which combines the $3\times3$ smoothing term, the compactness term, and the egde snapping.

The visual effect of these priors is illustrated in Fig.~\ref{impriors} and the impact of the priors on the performance is shown in Fig.~\ref{imevalpriors}. 
This experiment shows that the boundary priors have little impact on the undersegmentation error (CUE, UE and ASA), except when strictly enforcing compactness. 
The experiment also shows that all priors impact the boundary recall negatively. It seems that boundary recall is best when boundaries are allowed to update without the constraint of a prior.
Furthermore, the combined prior produces visually pleasing superpixels, and is a compromise between compact superpixels and good performance. 
However, if compact superpixels are not required, it seems advantageous to not enforce compactness at all.
For the remainder of the experiments no boundary prior term is used.

\begin{figure*}[t!]
\begin{tabular}{@{\hspace*{0.14cm}}c@{\hspace*{0.14cm}}c@{\hspace*{0.14cm}}c}
\includegraphics[width=0.32\textwidth]{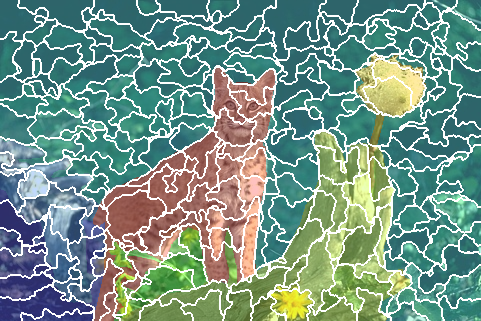} &
\includegraphics[width=0.32\textwidth]{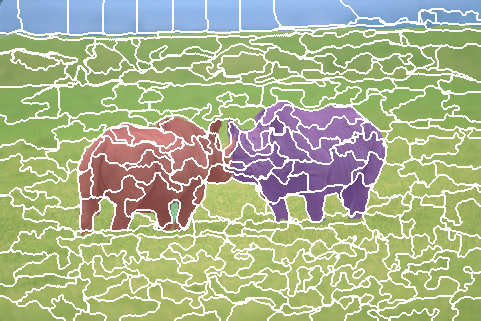}&
\includegraphics[width=0.32\textwidth]{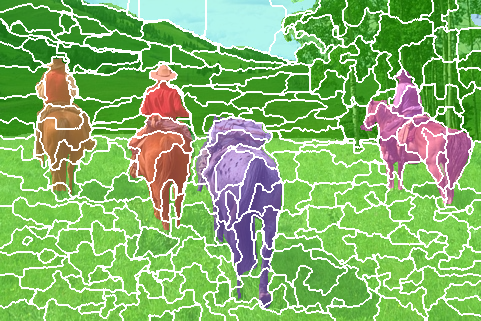}\\
\includegraphics[width=0.32\textwidth]{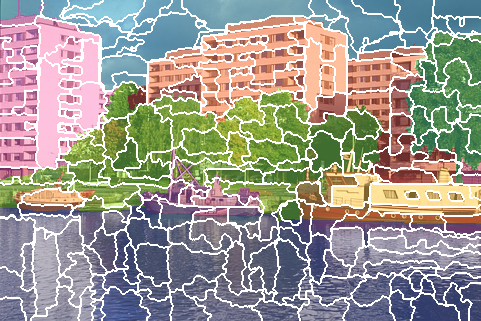}&
\includegraphics[width=0.32\textwidth]{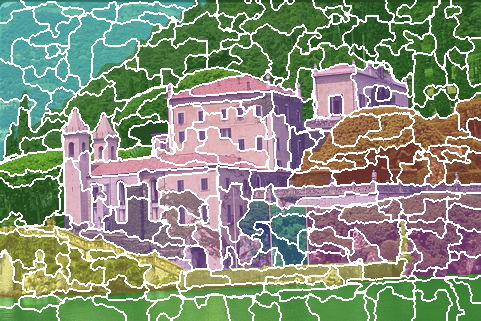}&
\includegraphics[width=0.32\textwidth]{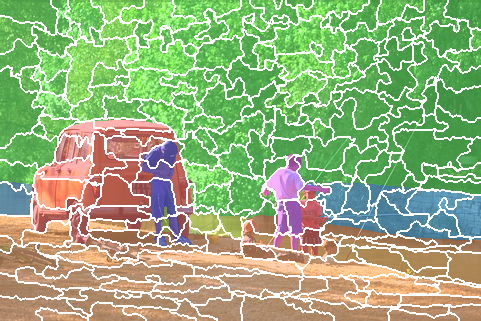}\\
\includegraphics[width=0.32\textwidth]{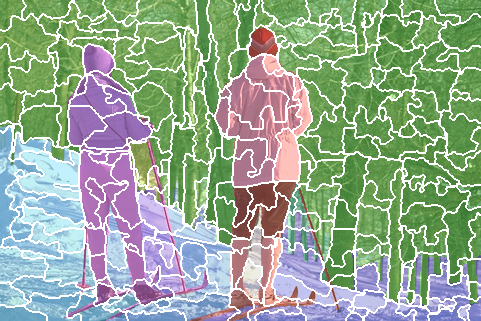}&
\includegraphics[width=0.32\textwidth]{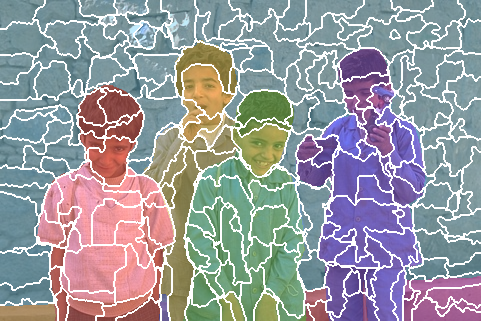}&
\includegraphics[width=0.32\textwidth]{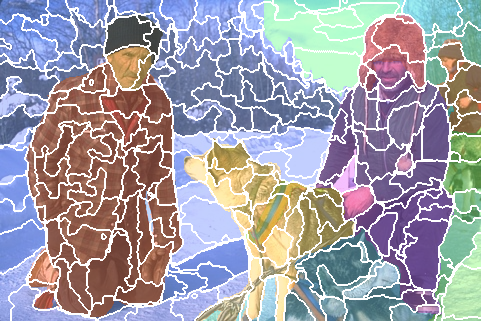}\\
\end{tabular}
\vspace{-0.2cm}
\caption{Example SEEDS segmentations with $200$ superpixels. The ground-truth segments are color coded and blended on the images. The superpixel boundaries are shown in white.}
\label{examples}
\end{figure*}

\subsection{Number of Iterations and Post Processing}

The hierarchical updating of the superpixel boundaries allows for a faster convergence of the SEEDS algorithm. A good segmentation can be obtained at 30 Hz, and the algorithm has enough time to converge in 15 Hz.

Fig.~\ref{imetimeBSD} shows that updating using means (SPM) converges significantly slower, but converges to a slightly better result. We propose to run this means-based updating as a post-processing step instead, in order to still benefit from that slight increase in performance. This is implemented by running the last few pixel-level updates based on means. Like this, we can combine the fast convergence of the histogram updating with the increased accuracy of the means-based updating. This is illustrated in Fig.~\ref{imevalpixel}.

\subsection{Comparison to State-of-the-Art}

We compare SEEDS to state-of-the-art methods Entropy Rate Superpixels\footnote{code available at \\ http://www.umiacs.umd.edu/$\sim$mingyliu/research}~\citep{Entropy11} (ERS),  
to SLIC\footnote{code available at \\ http://ivrg.epfl.ch/supplementary$\_$material/RK$\_$SLICSuperpixels}~\citep{SLIC10}, and 
to \cite{FH04} (FH)\footnote{code available at http://www.cs.brown.edu/$\sim$pff/segment/}.
 ERS is considered state-of-the-art in terms of performance, and 
SLIC is the fastest method available in the literature at 5 Hz. 
Note that, as FH does not output a fixed number of superpixels, the parameters are set such that the desired number of superpixels with the best performance were obtained.
We also show the performance of a plain grid (GRID) as a baseline to validate it as an initialization.

We report the results for two versions of SEEDS, one as presented in \cite{SEEDS} running at 5Hz, refered as SEEDS ECCV12. Another with the hierarchical updating proposed in this paper, 
refered as SEEDS, and runnning at 15Hz.
ERS ran at less than 1Hz in this experiment. 
The results (Fig.~\ref{imevalBSD}) show that SEEDS matches the UE and CUE, and outperforms the BR and ASA of ERS, while being orders of magnitude faster.

Additionally, in Fig.~\ref{figCD} we present results based on the BSDS300 contour detection benchmark~\citep{malik2011}, by running the superpixel algorithms as a contour detector. 
This is achieved by extracting superpixels on 12 different scales, ranging from $6$ to $600$ superpixels, and averaging the resulting boundaries. This is repeated for each superpixel algorithm.
SEEDS outperforms the other superpixel methods on this metric while being orders of magnitude faster.
Some examples of the segmentation results with 200 superpixels are shown in Fig.~\ref{examples}.

\section{Conclusions}

We have presented  a superpixel algorithm that achieves an excellent 
compromise between accuracy and efficiency. 
It is based on a hill-climbing optimization with efficient exchanges of pixels between superpixels.
The energy function that is maximized is based 
on enforcing homogeneity of the color distribution within superpixels.
The hill-climbing algorithm yields a very efficient evaluation of this energy function by using the intersection distance between histograms.
Its run-time can be controlled \emph{on the fly},
and we have shown the algorithm to run successfully 
in real-time, while staying competitive with the state-of-the-art on standard benchmark datasets. We use a single CPU and we do not use any GPU or dedicated hardware.

SEEDS performs well on the presented benchmarks, but we would also like to stress that it provides an extremely efficient framework for superpixels that can be adapted to many different applications. The energy function for updating the boundaries can be adapted to the application or the input sources. A variety of inputs can be used or combined, such as color, depth, optical flow, or video. The energy function can easily be adapted to take into account features other than color, such as texture or edges. All these adaptations are possible while maintaining all the real-time properties of the algorithm.

 The source code is available online\footnote{code available at http://www.vision.ee.ethz.ch/software}.

\begin{acknowledgements}
This work has been in part supported by the European Commission projects RADHAR (FP7 ICT 248873) and IURO (FP7 ICT 248314).
\end{acknowledgements}

\appendix
\section{ Evaluating Pixel-level and Block-level Movements}
\label{proofs}
In this section we prove both propositions used to speed up the evaluation of
the pixel-level and block-level movements.  

\subsection{Color Distribution Term}
Recall that $\mA^l_k$ is the set of pixels that are candidates to be moved from the 
superpixel $\mA_k$ to $\mA_n$.  

\paragraph{\bf{Proposition 1.}}\emph{
  Let the sizes of $\mA_k$ and $\mA_n$ be similar, and $\mA^l_k$ much smaller, \ie $|\mA_k|\approx |\mA_n| \gg |\mA_k^l|$.
If the histogram of $\mA^l_k$ is concentrated in a single bin, then
\begin{align}
\mathbf{int}(c_{\mA_n},c_{\mA_k^l}) \geq \mathbf{int}(c_{\mA_k \backslash \mA_k^l}, c_{\mA_k^l}) \Longleftrightarrow H(s) \geq H(s_t).
\end{align}}

\begin{proof}
 Recall that the color term of the energy function is:
\begin{align}
 H(s) = \sum_k \sum_{\{\mH_j\}} \left(  \frac{1}{|\mA_k|} \sum_{i \in \mA_k} \delta(I(i) \in \mH_j) \right)^2,
\end{align}
 in which we simply merged Eq.~\eqref{eqColorHistogram} and~\eqref{eqColorMain}.
 We write $H(s) \geq H(s_t)$ taking into account that $s$ and $s_t$ only differ in $\mA^l_k$, and the assumption of the 
Proposition on the size of the superpixels, \ie $|\mA_k|\approx |\mA_n| \gg |\mA_k^l|$. Thus, the
expression does not take into account the color at superpixels different from $k$ 
and $n$, and we can get rid of the normalization of the histograms due to the assumption. Then, the evaluation becomes,
\begin{align}
& H(s) \geq H(s_t) \Longleftrightarrow \nonumber\\
&\sum_{\{\mH_j\}} \left( \sum_{i \in \mA_n} \delta(I(i) \in \mH_j) + \sum_{i \in \mA_k^l} \delta(I(i) \in \mH_j) \right)^2 + \nonumber\\
&+ \sum_{\{\mH_j\}} \left( \sum_{i \in \mA_k\backslash \mA_k^l} \delta(I(i)\in \mH_j) \right)^2  \geq \nonumber\\
&\geq \sum_{\{\mH_j\}} \left( \sum_{i \in \mA_k\backslash \mA_k^l} \delta(I(i)\in \mH_j) + \sum_{i \in \mA_k^l} \delta(I(i)\in \mH_j )\right)^2 + \nonumber\\
& +\sum_{\{\mH_j\}} \left( \sum_{i \in \mA_n} \delta(I(i) \in \mH_j )\right)^2 .
\label{eq:proof1}
\end{align}

The second assumption of the Proposition is that $\mA^l_k$ is concentrated in a
 single bin. Let $\mH^*$ be the color in which $A^l_k$ is concentrated. Then, the evaluation in Eq.~\eqref{eq:proof1} becomes
\begin{align}
& \left( \sum_{i \in \mA_n} \delta(I(i) \in \mH^\star) + \sum_{i \in \mA_k^l} \delta(I(i) \in \mH^\star) \right)^2 +\nonumber\\
&+\sum_{\{\mH_j\} \backslash \mH^\star} \left( \sum_{i \in \mA_n} \delta(I(i) \in \mH_j) \right)^2 +\nonumber\\
&+\sum_{\{\mH_j\} } \left(\sum_{i \in \mA_k \backslash \mA_k^l} \delta(I(i) \in \mH_j) \right)^2 \geq \nonumber\\
&\geq \left( \sum_{i \in \mA_k \backslash \mA_k^l}  \delta(I(i) \in \mH^\star) + \sum_{i \in \mA_k^l} \delta(I(i)\in \mH^\star)\right)^2 +\nonumber\\
&+ \sum_{\{\mH_j\} \backslash \mH^\star} \left( \sum_{i \in \mA_k \backslash \mA_k^l} \delta(I(i) \in \mH_j) \right)^2+ \nonumber\\
&+ \sum_{\{\mH_j\}} \left( \sum_{i\in \mA_n} \delta(I(i) \in \mH_j)\right)^2.
\label{eq:proof2}
\end{align}
Then, note the following simple equality: 
\begin{align}
& \left( \sum_{i \in \mA_n} \delta(I(i) \in \mH^\star) + \sum_{i \in \mA_k^l} \delta(I(i) \in \mH^\star)\right)^2 =\\
&\left( \sum_{i \in \mA_n} \delta(I(i) \in \mH^\star)\right)^2 + \left(\sum_{i\in\mA_k^l} \delta(I(i)\in\mH^\star)\right)^2 + \nonumber\\
&+2 \left( \sum_{i\in\mA_n} \delta(I(i) \in \mH^\star) \right) \left( \sum_{i\in\mA_k^l} \delta(I(i) \in \mH^\star)\right),
\end{align}
 and we introduce it to the evaluation in Eq.~\eqref{eq:proof2}. Reordering the terms, and  canceling the same terms in both sides of the inequality, Eq.~\eqref{eq:proof2} becomes:
\begin{align}
& H(s) \geq H(s_t) \Longleftrightarrow \\
&\sum_{i\in\mA_n}\delta(I(i)\in\mH^\star) \geq \sum_{i\in\mA_k\backslash\mA_k^l}\delta(I(i)\in\mH^\star).
\label{eq:proof3}
\end{align}

Now, we develop the intersection distances in the Proposition to arrive to
Eq.~\eqref{eq:proof3}. We use the following expression:
\begin{align}
&\mathbf{int}(c_{\mA_n},c_{\mA_k^l}) =\\ 
&\sum_{\{\mH_j\}} \min \left( \frac{1}{|\mA_n|} \sum_{i\in\mA_n}\delta(I(i)\in\mH_j),\frac{1}{|\mA_k^l|} \sum_{i\in \mA_k^l} \delta(I(i) \in \mH_j) \right) \nonumber,
\end{align}
and since we assumed that the histogram of $\mA_k^l$ is concentrated in one bin, the expression becomes
\begin{align}
\mathbf{int}(c_{\mA_n},c_{\mA_k^l}) = \frac{1}{|\mA_n|}\sum_{i\in\mA_n}\delta(I(i)\in\mH^\star).
\end{align}
Finally, we use this expression and the assumption of $|\mA_k|\approx|\mA_n|$,and we obtain Eq.~\eqref{eq:proof3}:
\begin{align}
&\mathbf{int}(c_{\mA_n},c_{\mA_k^l}) \geq \mathbf{int}(c_{\mA_k\backslash\mA_k^l},c_{\mA_k^l}) \Longleftrightarrow \\
&\sum_{i\in\mA_n} \delta(I(i)\in\mH^\star) \geq \sum_{i \in \mA_k\backslash \mA_k^l}\delta(I(i)\in\mH^\star)  \Longleftrightarrow \\
&H(s)\geq H(s_t)
\end{align}\qed
\end{proof}

\subsection{Boundary Prior Term}

\paragraph{\bf{Proposition 2.}}
\emph{
Let $\{b_{\mN_i}(k)\}$ be the histograms of the superpixel labeling computed at the partitioning $s_t$ (see Eq.~\eqref{boundaryterm}). $\mA_k^l$ is a pixel, and $\mK_{\mA_k^l}$
 the set of pixels whose patch intersects with that pixel, \ie $\mK_{\mA_k^l} = \{i : \mA_k^l \in \mN_i \}$.
If the hill-climbing proposes moving a pixel $\mA_k^l$ from superpixel $k$ to superpixel $n$, then
\begin{align}
 \sum_{i\in\mK_{\mA_k^l}}(b_{\mN_i}(n)+1) \geq \sum_{i\in\mK_{\mA_k^l}} b_{\mN_i}(k) \Longleftrightarrow G(s) \geq G(s_t).
\end{align}}
\begin{proof}
Recall that $G(s)$ is:
\begin{align}
 G(s)=\sum_i \sum_k \left( \frac{1}{Z} \sum_{j \in \mN_i} \delta(j \in \mA_k)\right)^2,
\end{align}
where we merged Eq.~\eqref{boundaryterm} and~\eqref{eqBoundaryMain}. We write $G(s)\geq G(s_t)$ taking into account that $s$ and $s_t$ only differ in $\mA_k^l$, which is a single pixel, and it becomes
\begin{align}
& G(s) \geq G(s_t) \Longleftrightarrow \nonumber\\
&\sum_{i\in \mK_{\mA_k^l}} ( \left(\frac{1}{Z}((-1)+\sum_{j\in \mN_i}\delta(j \in \mA_k)) \right)^2 + \nonumber \\
&+\left(\frac{1}{Z}(1+\sum_{j\in\mN_i}\delta(j\in\mA_n)) \right)^2 ) \geq \nonumber \\
&\sum_{i \in \mK_{\mA_k^l}} \left( \left( \frac{1}{Z} \sum_{j\in\mN_i} \delta(j\in\mA_k) \right)^2 + \left( \frac{1}{Z} \sum_{j\in\mN_i}\delta(j\in\mA_n)\right)^2 \right).
\end{align}
    Then, we develop the squares, and cancel the repeated terms in the inequality
    as well as $Z$:
\begin{align}
& G(s) \geq G(s_t) \Longleftrightarrow  \nonumber \\
&\sum_{i\in\mK_{\mA_k^l}}  \left( 1-2 \sum_{j\in \mN_i}\delta(j\in\mA_k) \right) +  \nonumber \\
&+\left(1+2\sum_{j\in \mN_i}\delta(j\in\mA_n)\right) \geq 0.
\end{align}
    Finally, we reorder the terms and obtain the inequality in the Proposition: 
\begin{align}
&  G(s) \geq G(s_t) \Longleftrightarrow \nonumber \\
&\sum_{i\in\mK_{\mA_k^l}} \left(1+\sum_{j\in\mN_i}\delta(j\in\mA_n)\right) \geq \sum_{i\in\mK_{\mA_k^l}} \left( \sum_{j\in\mN_i} \delta(j\in\mA_k) \right)\Longleftrightarrow \nonumber \\
&\sum_{i\in\mK_{\mA_k^l}} (b_{\mN_i}(n)+1) \geq \sum_{i\in\mK_{\mA_k^l}} b_{\mN_i}(k).
\end{align}\qed
\end{proof}

\bibliographystyle{spbasic}      

\bibliography{egbib}

\end{document}